%% file: example.tex
\documentclass{article}

\usepackage[final]{corl_2026} 
\usepackage{times}

\usepackage[numbers]{natbib}

\usepackage{graphicx}
\usepackage{amsmath}
\usepackage{amssymb}
\usepackage{multirow}
\usepackage{booktabs}
\usepackage{tikz}
\usepackage{bbm}
\usepackage{bm}
\usepackage{pifont}
\usepackage{colortbl}
\usepackage{pifont}   
\usepackage{xcolor}   
\usepackage{caption}  
\usepackage{wrapfig}

\usepackage{float}
\usepackage{subcaption}

\newcommand{\cmark}{\textcolor{green!60!black}{\ding{51}}}
\newcommand{\xmark}{\textcolor{red!70!black}{\ding{55}}}

\title{MoRE: Mixture-of-Experts for Multi-Physics Robotic World Models}

\author{
Ziyu Zhu\textsuperscript{1,2,*},
Shaowei Liu\textsuperscript{1,*},
Albert J. Zhai\textsuperscript{1,*},
Henry Che\textsuperscript{1},
Wei-Chiu Ma\textsuperscript{3},
Shenlong Wang\textsuperscript{1}
\\[1ex]
\textsuperscript{1}University of Illinois Urbana-Champaign \quad
\textsuperscript{2}Georgia Institute of Technology 
 \quad
\textsuperscript{3}Cornell University
\\
\textsuperscript{*}Equal contribution.
}

\begin{document}
\maketitle

\input{figs/teaser}
\input{0_abstract}
\input{1_intro_v2}
\input{2_related_v2}
\input{3_method_v2}
\input{4_exp}

\input{5_conclusion}

\acknowledgments{This project is supported by the NSF Awards \#2331878, \#2340254, \#2312102, \#2414227, \#2404385, and \#2525287. We greatly appreciate the NCSA for providing computing resources through Delta and Delta AI program. We thank the support from the NVIDIA Academic Grant Program for Researchers.}


\bibliography{references}  

\input{supp}

\end{document}

%% file: figs/teaser.tex

\begin{figure*}[htbp]
  \centering
  \includegraphics[width=0.99\linewidth]{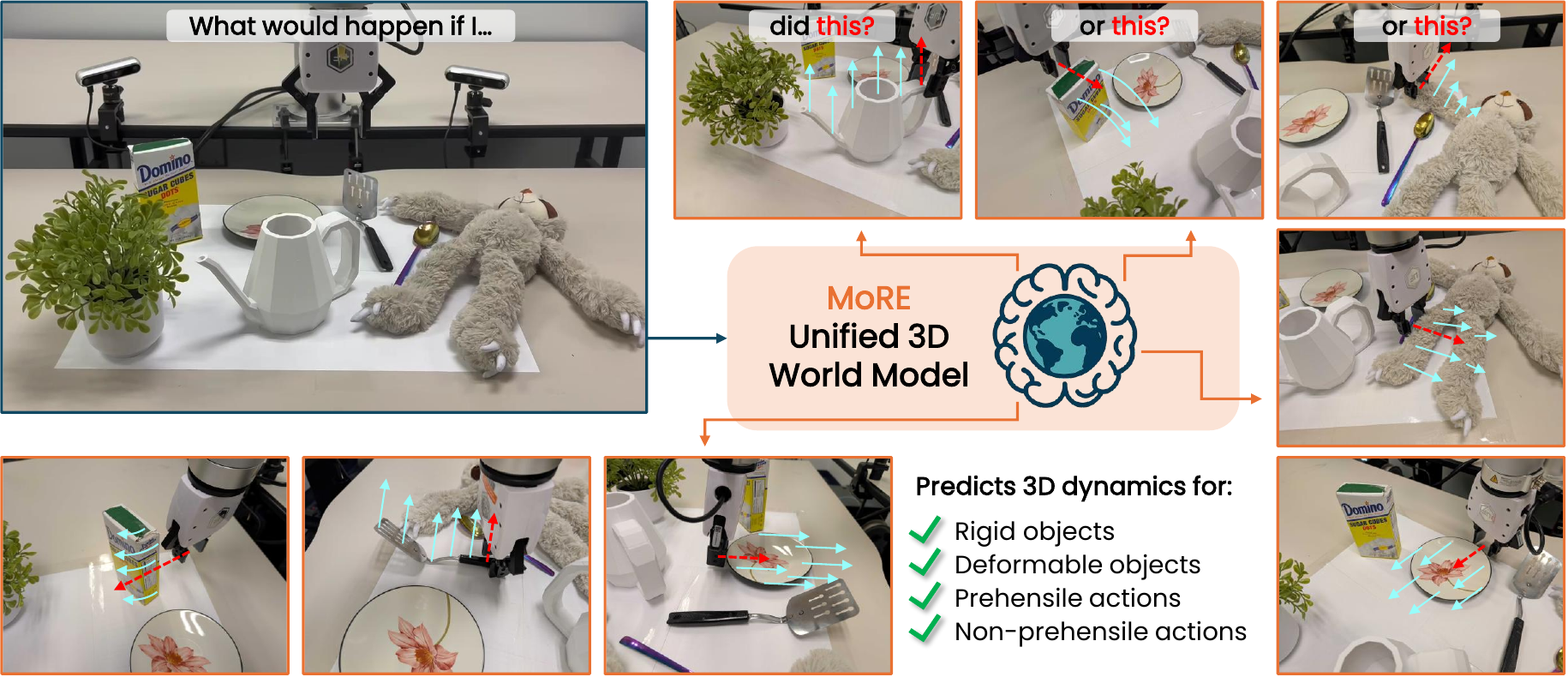}
  \caption{
\textbf{MoRE} predicts action-conditioned 3D dynamics by routing tool-object interactions to specialized motion experts. This unified MoE model handles both rigid and deformable objects, as well as both prehensile and non-prehensile contact. Trained on mixed simulation and real data, MoRE supports real-time prediction at 20 FPS and generalizes across tasks, materials, and objects.
}

\label{fig:teaser}
\end{figure*}

%% file: 0_abstract.tex
\begin{abstract}
A robotic world model should predict how objects evolve under robot actions and physical dynamics. While recent 3D world models have shown strong predictive capabilities, they still struggle with the heterogeneous dynamics arising from diverse object interactions. To address this in a unified framework, we decompose robotic interactions into distinct physical regimes and assign each to a specialized motion expert. We introduce MoRE, a 3D world model with a mixture of motion experts organized around rigid, deformable, and prehensile dynamics. This design enables a single model to capture diverse physical behaviors while preserving regime-specific priors. Trained on mixed simulation and real-world data, MoRE integrates with Model Predictive Control (MPC) to support diverse manipulation tasks. Experiments show that MoRE improves prediction accuracy across physics regimes, leading to stronger real-world MPC performance. Code, data, and project website are available at \href{https://alwaysleepy.github.io/MoRE-WM-site/}{MoRE-WM-site}.


\end{abstract}

%% file: 1_intro_v2.tex
\section{Introduction}


Robots that operate intelligently in the physical world should be able to predict how the environment evolves under their actions. Such an ability is fundamental for planning, control, and long-horizon decision making~\cite{zhou2024dino, zhou2024robodreamer, huang2026pointworld, ha2018world, openai2024video, guo2025ctrl, wu2023daydreamer}, and is especially crucial in contact-rich real-world manipulation~\cite{shi2023robocook, sumner2007embedded, lin2021softgym, zhang2025particle, zhang2024adaptigraph}. For instance, a house-cleaning robot must anticipate how cloth moves when wiping a surface, while a cooking robot must predict how ingredients deform under tool interactions.


\input{tables/paradim}

Building such a dynamics model is challenging because real-world manipulation involves heterogeneous physical behaviors, complex contacts, and diverse interaction mechanisms. To capture these effects, a robotic world model must be accurate, physically plausible, 3D-aware, and grounded in robot-object interactions. It should generalize across physical regimes, from rigid objects to deformable materials, and across manipulation modes, including prehensile grasping and non-prehensile interactions with rigid or soft tools. In short, we need models that are geometry-grounded, multi-physics, multi-tool, and contact-rich.


Unfortunately, existing approaches address only parts of this problem. Physics-based methods~\cite{jiang2025phystwin, xie2024physgaussian, du2021diffpd} leverage differentiable simulators and can support long-horizon rollouts, but they often require heavy system identification and expensive optimization. In contrast, learning-based dynamics models~\cite{zheng2025flare, chen2022comphy, li2018learning, lin2022learning, ma2023learning, xu2019densephysnet, battaglia2016interaction, sanchez2020learning, shi2024robocraft, zhang2025particle} avoid explicit system identification by learning dynamics directly from data, yet most are designed for a specific material type, tool, or task, limiting their ability to generalize across heterogeneous manipulation regimes. More recent generative world models~\cite{guo2025ctrl, hu2023gaia, zhou2024robodreamer, liu2026moright} offer greater scalability and broader scene coverage, but they are primarily optimized for visual prediction and often lack the accurate 3D structure and action grounding needed for real robotic execution. This leaves a gap between scalable prediction models and the physically grounded, controllable dynamics required for robot manipulation.


We aim to close this gap with MoRE, a scalable robotic world model that predicts future 3D point trajectories from a short history of object-tool interaction and the current robot action. Our key insight is that scaling alone is insufficient: to generalize across diverse robotic dynamics, scaling must be paired with a physics-informed mixture of motion experts. On the data side, we scale across simulation and real-world settings, covering diverse object geometries, materials, contact patterns, tasks, and tools. On the model side, we avoid forcing all dynamics into a single monolithic predictor. Instead, inspired by interaction-centric dynamics modeling~\cite{battaglia2016interaction, li2018learning, zhang2024adaptigraph, zhang2025particle, robine2023transformer} and Mixture-of-Experts (MoE) architectures~\cite{shazeer2017outrageously, fedus2022switch}, MoRE decomposes robotic dynamics into specialized motion experts aligned with physical regimes and interaction types, such as rigid/non-rigid and prehensile/non-prehensile. These experts share a common encoder for scalable representation learning, while each expert incorporates regime-specific inductive biases and constraints, such as rigidity preservation. A learned gating module routes each interaction to the appropriate expert based on short motion history, allowing the model to infer material properties and interaction modes directly from observation.

We evaluate MoRE in both simulation and real-world manipulation, covering pushing and grasping with rigid and deformable objects. Trained on a large-scale interaction dataset, our model runs at 20 FPS for model inference and integrates naturally with tuning-free MPC for closed-loop control. Across dynamics benchmarks and real robot tasks, MoRE consistently outperforms prior methods.


In summary, we make three contributions. First, we introduce MoRE, a 3D robotic world model for contact-rich dynamics prediction. Second, we design MoRE around a Mixture-of-Motion Experts architecture that explicitly captures heterogeneous physical regimes and interaction types. Third, we demonstrate that MoRE enables accurate real-time prediction and effective closed-loop control across both simulated and real-world robotic manipulation tasks.

%% file: tables/paradim.tex
\newcommand{\hmark}{\cmark$^\dagger$}

\begin{wraptable}{r}{0.52\textwidth}
\centering
\caption{\textbf{Comparison of robotic manipulation methods} across key capabilities: feed-forward inference, multi-physics dynamics (different material properties of manipulated objects, e.g., rigid vs. deformable), and multi-tool interaction (different end-effector geometries and contact interfaces), as well as 3D-awareness. ParticleFormer shows partial multi-tool support but is trained per scene. PointWorld relies on implicit dynamics while we explicitly route interactions to physics-aligned experts.}
\vspace{0.5em}
\label{tab:methodology_compare}
\setlength{\tabcolsep}{4pt}
\renewcommand{\arraystretch}{1}
\small
\begin{tabular}{lcccc}
\toprule
\textbf{Method}
& \begin{tabular}{@{}c@{}}\textbf{Feed-} \\ \textbf{forward}\end{tabular} 
& \begin{tabular}{@{}c@{}}\textbf{Multi-} \\ \textbf{phys.}\end{tabular} 
& \begin{tabular}{@{}c@{}}\textbf{Multi-} \\ \textbf{tool}\end{tabular} 
& \textbf{3D} \\
\midrule
PhysTwin~\cite{jiang2025phystwin} & \xmark & \xmark & \xmark & \cmark\\
PhysGaussian~\cite{xie2024physgaussian} & \xmark & \xmark & \xmark & \cmark \\
AdaptiGraph~\cite{zhang2024adaptigraph}      & \xmark & \xmark & \xmark & \cmark \\
ParticleFormer~\cite{huang2025particleformer}  & \cmark & \cmark & \hmark & \cmark \\
Points2Plans~\cite{huang2025points2plans}    & \cmark & \xmark & \xmark & \cmark \\
RoboDreamer~\cite{zhou2024robodreamer}     & \cmark & \xmark & \xmark & \xmark \\
DINO-WM~\cite{zhou2024dino}         & \cmark & \cmark & \cmark & \xmark \\
PointWorld~\cite{huang2026pointworld} & \cmark & \cmark & \cmark & \cmark \\
\textbf{Ours}              & \cmark & \cmark & \cmark & \cmark \\
\bottomrule
\end{tabular}

\end{wraptable}

%% file: 2_related_v2.tex
\section{Related Work}

Existing robotic dynamics models can be broadly divided into three categories: physics-based simulation methods, learning-based dynamics models, and foundation robotic world models. We compare our approach with representative prior work across these categories in Table~\ref{tab:methodology_compare}.

\paragraph{Physics-Based Dynamics Models}

Physics-based dynamics models~\cite{todorov2012mujoco, qiao2021differentiable, du2021diffpd} predict manipulation outcomes by explicitly simulating physical interactions. For rigid bodies~\cite{liu2024physgen, ferguson2021intersection, lan2022affine}, common formulations use state-based dynamics on the $\mathrm{SE}(3)$ manifold with contact constraints, implemented in rigid-body simulators~\cite{baraff1997introduction, baraff2001physically, todorov2012mujoco, coumans2021, commandeur2007introduction}. For deformable and soft objects, simulators typically rely on continuum or particle-based representations, such as MPM~\cite{xie2024physgaussian, chen2025physgen3d, de2020material, simo1998computational,hu2018moving, stomakhin2013material}, PBD~\cite{muller2007position, liu2023robotic, ai2025review}, or spring-mass systems~\cite{zhong2024reconstruction, jiang2025phystwin}, to capture material-dependent deformation. Recent hybrid pipelines~\cite{xie2024physgaussian, zhang2024physdreamer, chen2022virtual, le2023differentiable} combine reconstruction with physics-informed simulation to build digital twins for future prediction. These approaches provide strong physical priors and support long-horizon rollouts when parameters are accurate, but often depend on system identification and expensive test-time optimization to match observations~\cite{jiang2025phystwin, xie2024physgaussian}. Moreover, they typically require complete state and detailed geometry as input, making them suffer under partial observations, sensor noise, or unknown materials. 
In contrast, our method performs implicit system identification from interaction history, avoiding test-time optimization and explicit parameter identification.

\paragraph{Learning-Based Dynamics Models}
Learning-based dynamics models~\cite{zheng2025flare, chen2022comphy, li2018learning, lin2022learning, ma2023learning, xu2019densephysnet} bypass explicit simulation by learning action-conditioned state transitions from data. Early work represents objects and interactions using graphs~\cite{li2018learning, pfaff2020learning}, particles~\cite{zhang2025particle, guan2022neurofluid, li2018learning}, or GNNs~\cite{battaglia2016interaction, sanchez2020learning, shi2024robocraft}, while more recent approaches adopt transformer-style architectures to improve capacity and scalability for contact-rich interactions. These models enable efficient inference and strong performance in specific regimes, operating directly from observations. However, those models remain \emph{narrow}: they are trained on category or task-specific datasets and implicitly assume fixed materials, tools, or interaction modes. This often leads to overfitting and limited generalization ability. Our model is also learning-based, but is explicitly designed to support diverse materials, physical properties, object geometries, and action-conditioned interactions within a unified framework.

\paragraph{Foundation Robotic World Models}

Motivated by the success of large visual foundation models~\cite{openai2024video, bardes2024revisiting, oquab2023dinov2}, recent robotic world models can be broadly grouped into 2D and 3D approaches. 2D methods~\cite{guo2025ctrl, robine2023transformer, hansen2023td, mendonca2023structured, hu2023gaia} leverage pretrained visual features or video generative models for prediction and planning; for example, DINO-WM~\cite{zhou2024dino} plans in a DINOv2~\cite{oquab2023dinov2} feature space, while RoboDreamer~\cite{zhou2024robodreamer} synthesizes future videos with diffusion models~\cite{ko2023learning}. However, these methods are not explicitly grounded in 3D geometry or contact interactions. 3D methods instead predict scene evolution in geometric space. PointWorld~\cite{huang2026pointworld}, for example, predicts action-conditioned 3D point flow from scene point clouds and robot motion. While promising, existing 3D approaches do not explicitly model the heterogeneous dynamics of robotic interactions. In contrast, our method uses learned gating to route interactions to regime-specific motion experts, enabling more physically faithful prediction across diverse manipulation scenarios.


%% file: 3_method_v2.tex
\section{Method}
\input{figs/method}

\subsection{Problem Formulation}

We model a robotic \emph{3D world model} as a dynamical system that predicts future scene evolution under robot actions and physical interactions.
Let $\mathbf{s}_t \in \mathcal{S}$ denote the \emph{current} scene state and $\mathbf{a}_t \in \mathcal{A}$ the \emph{current} robot action,
where $\mathcal{S}$ and $\mathcal{A}$ are geometry-based point-cloud spaces defined below.
Given the current state--action pair $(\mathbf{s}_t, \mathbf{a}_t)$ together with a finite history of past states and actions
$\{\mathbf{s}_{t-H':t-1}, \mathbf{a}_{t-H':t-1}\}$, the goal is to predict future scene states over a horizon of $H$ steps.

We parameterize the world model as
\begin{equation}
\label{eq:world_model}
\mathbf{s}_{t+1:t+H}
=
f_\theta
\big(
\mathbf{s}_t,\,
\mathbf{a}_t,\,
\mathbf{s}_{t-H':t-1},\,
\mathbf{a}_{t-H':t-1}
\big).
\end{equation}
We represent each scene state as a 3D point cloud,
$\mathbf{s}_t \in \mathcal{S} := \mathbb{R}^{N_S \times 3}$ with $N_S$ being the number of scene points.
Crucially, we design the action space $\mathcal{A}$ in a geometry-centric manner:
unlike most prior work that conditions dynamics on joint commands, end-effector poses, or sparse keypoints~\cite{levine2016end, shridhar2023perceiver, di2024keypoint},
we represent each action by the full 3D state of the \emph{tool} as a point cloud,
$\mathbf{a}_t \in \mathcal{A} := \mathbb{R}^{N_A \times 3}$.
Specifically, low-level robot controls are mapped via forward kinematics to $\mathbf{a}_t$,
where the tool is any directly actuated geometry by the embodiment
(e.g., grippers, hands, cookware, or daily tools), as long as its state can be computed via forward kinematics.
This representation decouples the world model from embodiment-specific kinematics, enabling cross-embodiment transfer
and multi-tool generalization across diverse manipulation tasks.
Conditioned on the current and historical scene--action point clouds, the model predicts future scene point clouds
$\mathbf{s}_{t+1:t+H}$, capturing rigid-body motion, deformation, and contact-induced dynamics.

\subsection{Mixture-of-Experts Dynamic Modeling}
\label{sec:moe_dynamics}
Real-world physical interactions span diverse regimes, from rigid to non-rigid dynamics and from prehensile to non-prehensile manipulation.
These regimes impose distinct kinematic and kinetic constraints, making a single monolithic transformer difficult to learn and optimize.
Yet, we observe that the interaction regime is often straightforward to infer from a short history of states and actions. Motivated by this, we adopt a Mixture-of-Experts architecture that identifies motion patterns and routes to specialized dynamics decoders for multi-physics modeling (Fig.~\ref{fig:framework}).

Given the current and historical scene states and actions represented as point clouds, we encode them using a scene encoder and an action encoder to obtain point \emph{embeddings}.
The embeddings are fused into a unified representation $\mathbf{z}_t$, serving as shared conditioning for both the gating function and the expert dynamics models.
We predict a one-hot routing vector via a gating function $g(\cdot)$:
\begin{equation}
\mathbf{g}_t = g_\theta(\mathbf{z}_t);\text{\ where \ }
\mathbf{g}_t \in \{0,1\}^{K},
\sum_{k=1}^{K} g_t^{(k)} = 1,
\end{equation}
where $K$ is the number of experts. The router is re-evaluated at each rollout step, allowing adaptation to changes in the interaction regime.
This hard routing encourages clear expert specialization and reduces interference across heterogeneous physical regimes.

Each expert dynamics model $f_k$ takes the same shared embedding $\mathbf{z}_t$ as input and predicts the next H-step scene states.
The overall MoE prediction is $\mathbf{s}_{t+1:t+H}
=
\sum_{k=1}^{K} g_t^{(k)}\, f_k(\mathbf{z}_t),$
which is equivalent to activating only the selected expert under one-hot routing.

\subsection{Physics-Aware Expert Design}
\label{sec:expert_dynamics}


We group the experts into two material classes, rigid and deformable, and further specialize them by interaction regime. 
Our model uses three experts: (i) \emph{rigid non-prehensile}, (ii) \emph{deformable non-prehensile}, and (iii) \emph{deformable prehensile}. 
Each expert is first trained on its corresponding simulation-domain data and then jointly fine-tuned on mixed real and simulated data. 
We omit rigid prehensile interactions, since rigid grasping has limited physical uncertainty when slip is negligible and can be well modeled by existing world models.

\paragraph{Rigid Expert.}
For rigid objects, motion is fully characterized by a global 6-DoF rigid transform $(\mathbf{R}_\tau,\mathbf{t}_\tau)\in \mathrm{SE}(3)$.
Accordingly, the rigid expert predicts a sequence of transforms via an internal head
$
\left(\boldsymbol{\omega}_{t+1:t+H}, \mathbf{t}_{t+1:t+H}\right)
=
h_{\text{rigid}}(\mathbf{z}_t),
$
where $\boldsymbol{\omega}_t \in \mathbb{R}^6$ is the 6D continuous rotation representation (converted to $\mathbf{R}_t\in\mathrm{SO}(3)$) and $\mathbf{t}_t \in \mathbb{R}^3$ is translation.
The expert then outputs future scene point clouds by rigid warping
$f_{\text{rigid}}(\mathbf{z}_t;\tau)
=
\mathbf{R}_{t+1:t+H} \cdot \mathbf{s}_{t} + \mathbf{t}_{t+1:t+H}
$
This parameterization enforces rigidity and global coherence, reducing the hypothesis space and improving stability compared to free-form point decoding.

\paragraph{Deformable Experts.}
Deformable objects require free-form geometry updates beyond global rigid warping.
We therefore instantiate \emph{two} deformable experts with identical architecture but separate parameters and training: one specialized for prehensile interactions and one for non-prehensile interactions. The prehensile expert specializes in force-induced shape changes under grasping (e.g., squeezing and stretching),
whereas the non-prehensile expert focuses on contact-induced deformations from pushing and sliding interactions.
For simplicity, we omit explicit expert indices in the notation. Given the shared embedding $\mathbf{z}_t$, the selected deformable expert predicts point-wise $
\Delta \mathbf{s}_{t+1:t+H}
=
h_{\text{def}}(\mathbf{z}_t),
$
and outputs future scene states by applying the displacement
$
f_{\text{def}}(\mathbf{z}_t)
=
\mathbf{s}_{t:t+H-1} + \Delta \mathbf{s}_{t+1:t+H}.
$

\subsection{Training and Inference}

\paragraph{Training}
During training, we adopt a self-rolling strategy~\cite{huang2025self} to match inference-time behavior and mitigate distribution shift.
Starting from the current scene state $\mathbf{s}_t$ (and the history buffer), the model predicts the next-step scene $\hat{\mathbf{s}}_{t+1}$.
We then feed $\hat{\mathbf{s}}_{t+1}$ back as input to predict subsequent steps, rolling out autoregressively over a horizon of $H$ steps.
Following prior work, gradients are detached through the rolled-in predictions, and losses are computed over the entire rollout horizon.

We optimize a combined objective consisting of a reconstruction loss and an expert routing loss:
\begin{equation}
\label{eq:train_loss}
\mathcal{L}
=
\sum_{\ell=1}^{H}
\left\|
{\mathbf{s}}_{t+\ell}
-
\mathbf{s}^{\text{gt}}_{t+\ell}
\right\|_2^2
+
\lambda \sum_{\ell=1}^{H}
\mathrm{CE}\!\left({\mathbf{g}}_{t+\ell},\, \mathbf{g}^{\text{gt}}_{t+\ell}\right),
\end{equation}
where $\mathbf{s}^{\text{gt}}$ denotes ground-truth scene point clouds,
$\hat{\mathbf{g}}$ is the predicted one-hot routing distribution from the gating function $g(\cdot)$,
and $\mathbf{g}^{\text{gt}}$ is the supervision signal for expert selection.
We first warm up the gating network and experts separately on simulation data, and then jointly fine-tune the full model on mixed real-world data.

\paragraph{Inference}
At inference time, the model performs fully autoregressive rollout.
Given the observed initial scene and a history buffer initialized from past states/actions, we iteratively predict ${\mathbf{s}}_{t+1}, {\mathbf{s}}_{t+2}, \ldots$.
After each step, the predicted state is appended to the history buffer and used as input for the next prediction.
The procedure repeats until reaching the desired prediction horizon.

\subsection{Planning with World Model}
We use the learned world model as the predictive dynamics in an Model Predictive Control (MPC) loop.
At each control step, given the current observed state $s_t$ (scene point cloud) and history, we sample a set of candidate action sequences $\{\mathbf{a}_{t:t+H-1}^{(j)}\}_{j=1}^{J}$ over a planning horizon $H$ and roll out the world model to obtain predicted trajectories $\{{\mathbf{s}}_{t+1:t+H}^{(j)}\}$.
We evaluate each trajectory with a task-specific cost that measures progress toward a goal state $s_{\text{goal}}$ (e.g., distance between the predicted terminal scene and the goal), and select the lowest-cost plan $j^\star$.
We then execute only the first action $a_t^{(j^\star)}$, observe the next state, and re-plan at the next step. For obstacle present (e.g. multi-object manipulation), we add additional cost for collision avoidance. 
Other MPC details (action sampling, costs, and constraints) are provided in the following sections.

%% file: figs/method.tex
\begin{figure*}[t]
  \centering
  \includegraphics[width=0.99\linewidth]{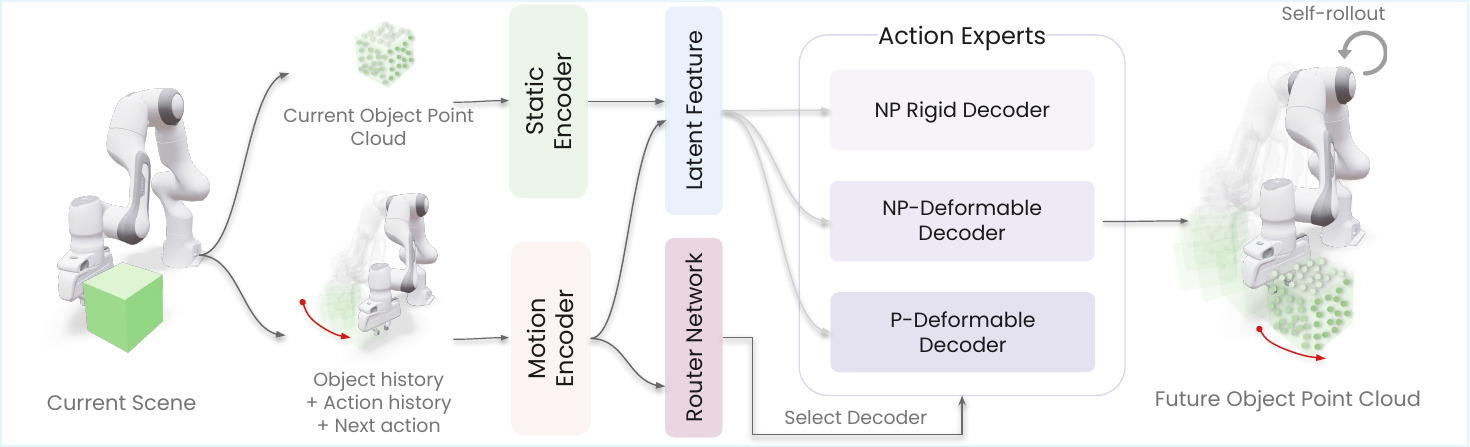}
  \caption{\textbf{Overview of MoRE}. Given the current and historical object states and actions, represented as point clouds, we first encode the scene and actor inputs into point tokens using two encoders. An explicit gating network then selects among specialized motion experts, each tailored to a distinct physical regime, to predict the next object state in 3D. P = prehensile, NP = non-prehensile.}
  \label{fig:framework}

\end{figure*}

%% file: 4_exp.tex
\input{tables/sim_comparison}
\input{tables/sim_ablation}

\input{tables/droid}
\input{figs/droid}

\input{figs/mpc}
\input{figs/real}

\section{Experiments}

\subsection{Implementation Details}

\paragraph{Dataset}

We use a mixed simulation-real dataset for training, covering diverse physical regimes, and interaction types.
The simulation data are collected using MagicSim~\cite{lu2026magicsimunifiedinfrastructureexecutable} from Isaac Sim and
contain approximately 20K episodes.
We automatically generate diverse interactions with random pushing actions, and physical properties are randomized per episode. The real-world data contain approximately 7K episodes of DROID~\cite{khazatsky2024droid} and~\cite{zhang2024adaptigraph}. Additional dataset details can be found in the supplementary.

\paragraph{Model Architecture}
Our model takes a temporal sequence of point sets as input, using 8 history steps to predict a 3-step horizon. The object point cloud is represented with a variable number of points based on object size, together with 100 end-effector points. Details about network architecture can be found in the supplementary.


\subsection{Tasks and Evaluation Setup}

\paragraph{Tasks}
Our evaluation covers three representative manipulation regimes. In rigid non-prehensile tasks, 30 different tools from~\cite{lin2025robotsmith} are used to push or pull 300 rigid objects from~\cite{chen2025robotwin20scalabledata}, emphasizing object coherence and contact consistency. In deformable non-prehensile tasks, the same tools interact with 300 soft objects from~\cite{stojanov21cvpr} through pushing and pulling, requiring accurate modeling of local deformation. In deformable prehensile tasks, the robot grasps and moves deformable objects such as toys and clothes, introducing sustained contact and large non-rigid shape changes.


\paragraph{Metrics and Evaluation Setup}
We evaluate dynamic prediction with geometric accuracy and material-consistent structural metrics, as summarized in Table~\ref{tab:dynamics_comparison}. 
Geometric accuracy is measured by MSE and Chamfer Distance (CD) between predicted and ground-truth point clouds. 
For rigid interactions, we report rigidity error by fitting an optimal rigid transformation with ICP and measuring residual alignment error. 
For deformable objects, we report Laplacian error between predicted and ground-truth point clouds, where lower values indicate smoother, more plausible deformations.

\subsection{Dynamic Modeling across Interactions and Physics}

We compare against different baselines in Tables~\ref{tab:dynamics_comparison} and~\ref{tab:droid}. 
In Table~\ref{tab:dynamics_comparison}, we compare with AdaptiGraph~\cite{zhang2024adaptigraph} and ParticleFormer~\cite{huang2025particleformer}, which are neural dynamics models mainly designed for deformable object modeling, as well as privileged experts trained on specific interaction domains.

Across all regimes in Table~\ref{tab:dynamics_comparison}, existing methods perform well only within their targeted categories and struggle to generalize across interaction types and physical properties. 
In contrast, our MoE model does not rely on privileged test-time interaction labels, yet achieves performance comparable to domain-specific experts across diverse physics regimes. 
Its shared representation and routing enable effective specialization for heterogeneous manipulation dynamics. 
We additionally compare with DINO-WM~\cite{zhou2024dino}, a 2D feature-based world model. Since point-wise RMSE is not directly comparable, we evaluate Chamfer Distance, where MoRE achieves 0.048 compared with 0.071 for DINO-WM. We also perform 10 MPC evaluations in simulation with diverse objects and end-effectors. MoRE achieves a success rate of 50\%, compared with 20\% for DINO-WM. These results indicate that MoRE achieves better dynamics prediction accuracy and supports more effective downstream control than DINO-WM. The relatively low absolute MPC success rates reflect the challenging evaluation setting, where errors can accumulate during long-horizon rollouts.

In Table~\ref{tab:droid}, we further compare with PointWorld on DROID, using its model trained directly on DROID data. 
Since our goal is manipulated-object dynamics, we evaluate only moving object points rather than whole-scene flow. 
Our method achieves better performance across all three cases, demonstrating strong generalization to real-world object dynamics. Visualizations of different scenes and comparisons between our predictions and ground truth are shown in Fig.~\ref{fig:droid}.

\subsection{Ablation Study on Motion Experts and Routing}
Table~\ref{tab:dynamics_ablation} presents an ablation study on architectural and training design choices. 
All ablation models are trained on the same dataset with approximately matched model capacity.
Training a single expert to model all physics leads to degraded performance, struggling to capture both rigid and deformable behaviors. Implicit routing, which softly combines expert outputs without supervised binary gating, further reduces geometric accuracy and physical consistency, highlighting the importance of explicit expert selection. Implicit MoE embeds expert specialization within the transformer using a shared, displacement-based output representation, rather than employing explicit motion experts with distinct roles. This design consistently underperforms across all regimes, suggesting that implicit capacity allocation alone is insufficient to capture heterogeneous dynamics. Finally, removing self-rollout training and relying solely on teacher forcing results in less stable predictions, particularly for contact-rich deformable interactions. 

\subsection{MPC Performance with Learned World Models}
\paragraph{Setup}
We evaluate downstream control performance with MPC in real-world robot experiments, including rigid box pushing and deformable toy-sloth pose matching. 
Goals are specified as target object point clouds, and the MPC cost is defined as the bidirectional Chamfer Distance (CD) between predicted and target point clouds. 
Success is measured using a threshold on the actual CD after execution. 
We compare against two baselines: \emph{Heuristic MPC}, which assumes all object points follow the end-effector translation, and \emph{Vanilla*}, a single Transformer-based world model without motion experts or routing. 
Additional MPC details are provided in the supplementary.

\paragraph{Results}
As shown in Fig.~\ref{fig:task_mpc_real}, our method consistently achieves lower CD compared to baselines. Heuristic MPC struggles to generalize across tasks due to hand-designed, task-specific objectives, while Vanilla* suffers from limited capacity to model heterogeneous physical dynamics. In contrast, our physics-aware, expert-based 3D world model enables more accurate long-horizon prediction, translating to more efficient MPC. Prediction rollouts are visualized in Fig.~\ref{fig:real_preds}. 

%% file: tables/sim_comparison.tex
\begin{table*}[t]
\centering
\caption{
\textbf{Dynamic model comparison} across interaction types and physics. We report geometric accuracy (MSE, CD) and material structural metrics (rigidity or Laplacian). Privileged experts (gray) are domain-specific and assume known interaction types at test time, while our Mixture-of-Experts model handles diverse interactions in a unified framework with consistently strong performance.}
\label{tab:dynamics_comparison}
\setlength{\tabcolsep}{2pt}
\renewcommand{\arraystretch}{1.2}
\small
\begin{tabular}{lccc|ccc|ccc}
\toprule
 & \multicolumn{3}{c}{\textbf{\shortstack{Rigid-body\\Non-Prehensile}}}
 & \multicolumn{3}{c}{\textbf{\shortstack{Deformable\\Non-Prehensile}}}
 & \multicolumn{3}{c}{\textbf{\shortstack{Deformable\\Prehensile}}} \\
\cmidrule(lr){2-4} \cmidrule(lr){5-7} \cmidrule(lr){8-10}
\textbf{Method}
 & \textbf{RMSE} $\downarrow$ 
 & \textbf{CD} $\downarrow$ 
 & \textbf{Rigidity} $\downarrow$
 & \textbf{RMSE} $\downarrow$ 
 & \textbf{CD} $\downarrow$ 
 & \textbf{Laplacian} $\downarrow$
 & \textbf{RMSE} $\downarrow$ 
 & \textbf{CD} $\downarrow$ 
 & \textbf{Laplacian} $\downarrow$ \\
\midrule
AdaptiGraph         & 0.028 & 0.084 & 0.0001 & 0.031 & 0.052 & 0.0068 & 0.047 & 0.143 & 0.0092 \\
ParticleFormer     & 0.054 & 0.082 & 0.0005 & 0.061 & 0.079 & 0.0072 & 0.058 & 0.075 & \underline{0.0058} \\
\midrule
\rowcolor{gray!10}
Privileged Expert  & \textbf{0.021} & \textbf{0.048} & \textbf{0.0} & \textbf{0.018} & \underline{0.032} & \underline{0.0027} & 0.029 & \textbf{0.038} & 0.0068 \\

\midrule
\textbf{Ours (MoE)} 
& \underline{0.022} & \textbf{0.048} & \textbf{0.0} 
                  & \underline{0.024} & \textbf{0.027} & \textbf{0.0020} 
                  & \textbf{0.026} & \underline{0.039} & \textbf{0.0050} \\
\bottomrule
\end{tabular}
\end{table*}

%% file: tables/sim_ablation.tex
\begin{table*}[t]
\centering
\caption{\textbf{Ablation of MoRE} under different interaction types and physical regimes. We evaluate the impact of expert specialization, routing strategy, and self-rollout training using geometric and structural metrics. Performance degrades when removing expert specialization, explicit routing, or self-rollout training, highlighting the importance of each component for robust dynamics prediction.}
\label{tab:dynamics_ablation}
\setlength{\tabcolsep}{2pt}
\renewcommand{\arraystretch}{1.2}
\small
\begin{tabular}{lccc|ccc|ccc}
\toprule
 & \multicolumn{3}{c}{\textbf{\shortstack{Rigid-body\\Non-Prehensile}}}
 & \multicolumn{3}{c}{\textbf{\shortstack{Deformable\\Non-Prehensile}}}
 & \multicolumn{3}{c}{\textbf{\shortstack{Deformable\\Prehensile}}} \\
\cmidrule(lr){2-4} \cmidrule(lr){5-7} \cmidrule(lr){8-10}
\textbf{Method}
 & \textbf{RMSE} $\downarrow$
 & \textbf{CD} $\downarrow$
 & \textbf{Rigidity} $\downarrow$
 & \textbf{RMSE} $\downarrow$
 & \textbf{CD} $\downarrow$
 & \textbf{Laplacian} $\downarrow$
 & \textbf{RMSE} $\downarrow$
 & \textbf{CD} $\downarrow$
 & \textbf{Laplacian} $\downarrow$ \\
\midrule
Single Expert     & 0.030 & 0.054 & 0.0001 & 0.028 & 0.038 & 0.0038 & 0.037 & 0.047 & 0.0061 \\
Implicit Router   & 0.026 & 0.051 & \textbf{0.0}    & 0.032 & 0.033 & 0.0032 & 0.028 & 0.040 & 0.0054 \\
Implicit MoE      & 0.045 & 0.070 & 0.0002 & 0.026 & 0.031 & 0.0032 & 0.032 & \textbf{0.039} & 0.0071 \\
w/o Self-Rollout  & 0.031 & 0.057 & \textbf{0.0}    & 0.035 & 0.039 & 0.0056 & 0.047 & 0.060 & 0.0058 \\
\midrule
\textbf{Ours (Full)} 
                  & \textbf{0.022} & \textbf{0.048} & \textbf{0.0} 
                  & \textbf{0.024} & \textbf{0.027} & \textbf{0.0020} 
                  & \textbf{0.026} & \textbf{0.039}  & \textbf{0.0050} \\
\bottomrule
\end{tabular}

\end{table*}

%% file: tables/droid.tex
\begin{table*}[t]
\centering
\caption{\textbf{Comparison with Point-World on the DROID dataset.} Our method performs better across all three cases, showing strong generalization to real-world object dynamics.}
\label{tab:droid}
\setlength{\tabcolsep}{2pt}
\renewcommand{\arraystretch}{1.2}
\small
\begin{tabular}{lccc|ccc|ccc}
\toprule
 & \multicolumn{3}{c}{\textbf{\shortstack{Rigid-body\\Non-Prehensile}}}
 & \multicolumn{3}{c}{\textbf{\shortstack{Deformable\\Non-Prehensile}}}
 & \multicolumn{3}{c}{\textbf{\shortstack{Deformable\\Prehensile}}} \\
\cmidrule(lr){2-4} \cmidrule(lr){5-7} \cmidrule(lr){8-10}
\textbf{Method}
 & \textbf{RMSE} $\downarrow$
 & \textbf{CD} $\downarrow$
 & \textbf{Rigidity} $\downarrow$
 & \textbf{RMSE} $\downarrow$
 & \textbf{CD} $\downarrow$
 & \textbf{Laplacian} $\downarrow$
 & \textbf{RMSE} $\downarrow$
 & \textbf{CD} $\downarrow$
 & \textbf{Laplacian} $\downarrow$ \\
\midrule
Point-world     & 0.031 & 0.054 & 0.0003 & 0.028 & 0.031 & 0.0041 & 0.029 & 0.035 & 0.0063 \\
\textbf{Ours} 
                  & \textbf{0.023} & \textbf{0.046} & \textbf{0.0} 
                  & \textbf{0.014} & \textbf{0.026} & \textbf{0.0022} 
                  & \textbf{0.021} & \textbf{0.031}  & \textbf{0.0051} \\
\bottomrule
\end{tabular}

\end{table*}

%% file: figs/droid.tex
\begin{figure*}[t]
  \centering
  \includegraphics[width=0.99\linewidth]{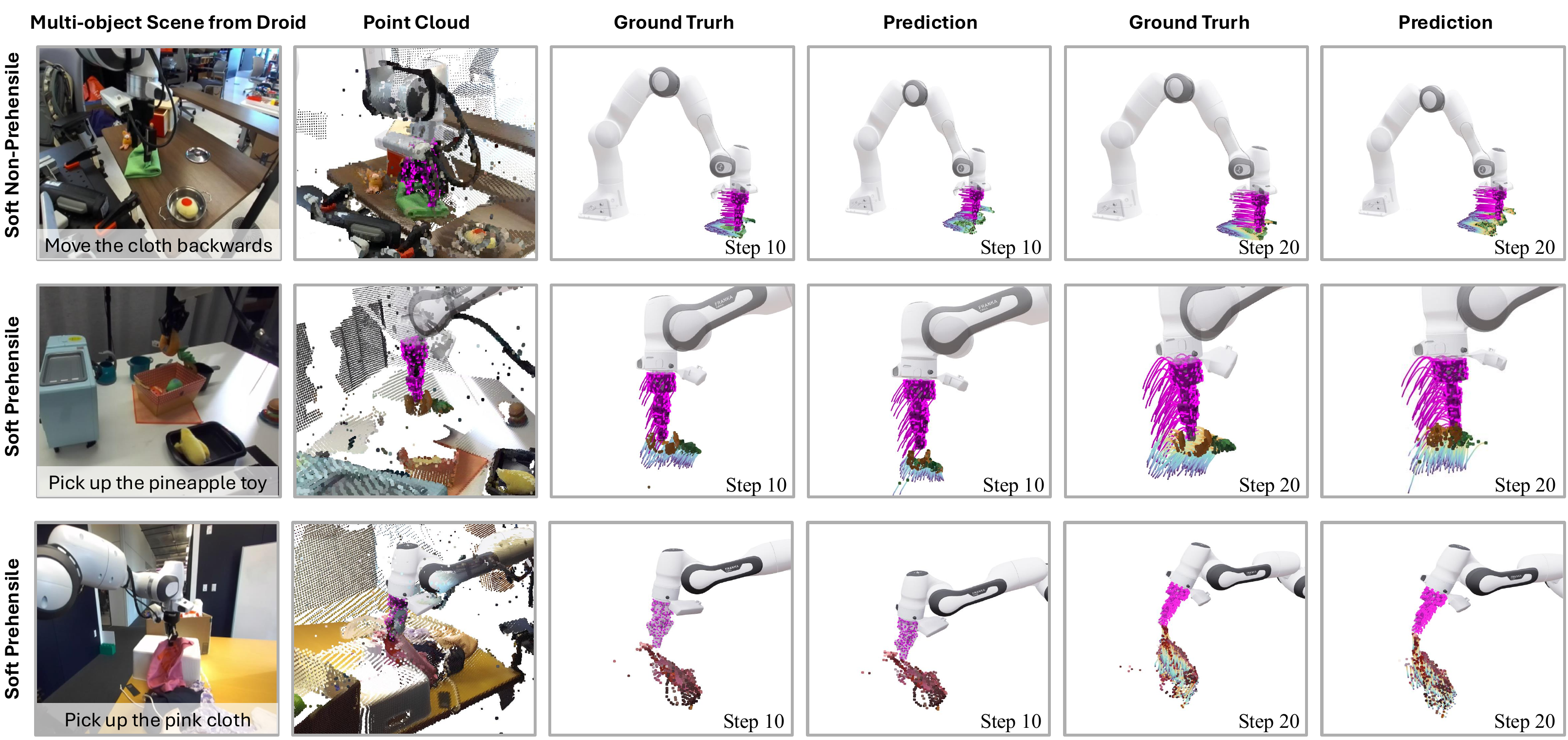}
  \caption{\textbf{Qualitative results on DROID evaluation.} We compare model predictions with ground truth on challenging DROID scenes with clutter, diverse objects, and partial observations. Even for small, low-resolution manipulated objects, our model produces accurate dynamics, demonstrating robustness to occlusion and noise.}
  \label{fig:droid}

\end{figure*}

%% file: figs/mpc.tex
\begin{figure*}[t]
  \centering
  \includegraphics[width=\linewidth]{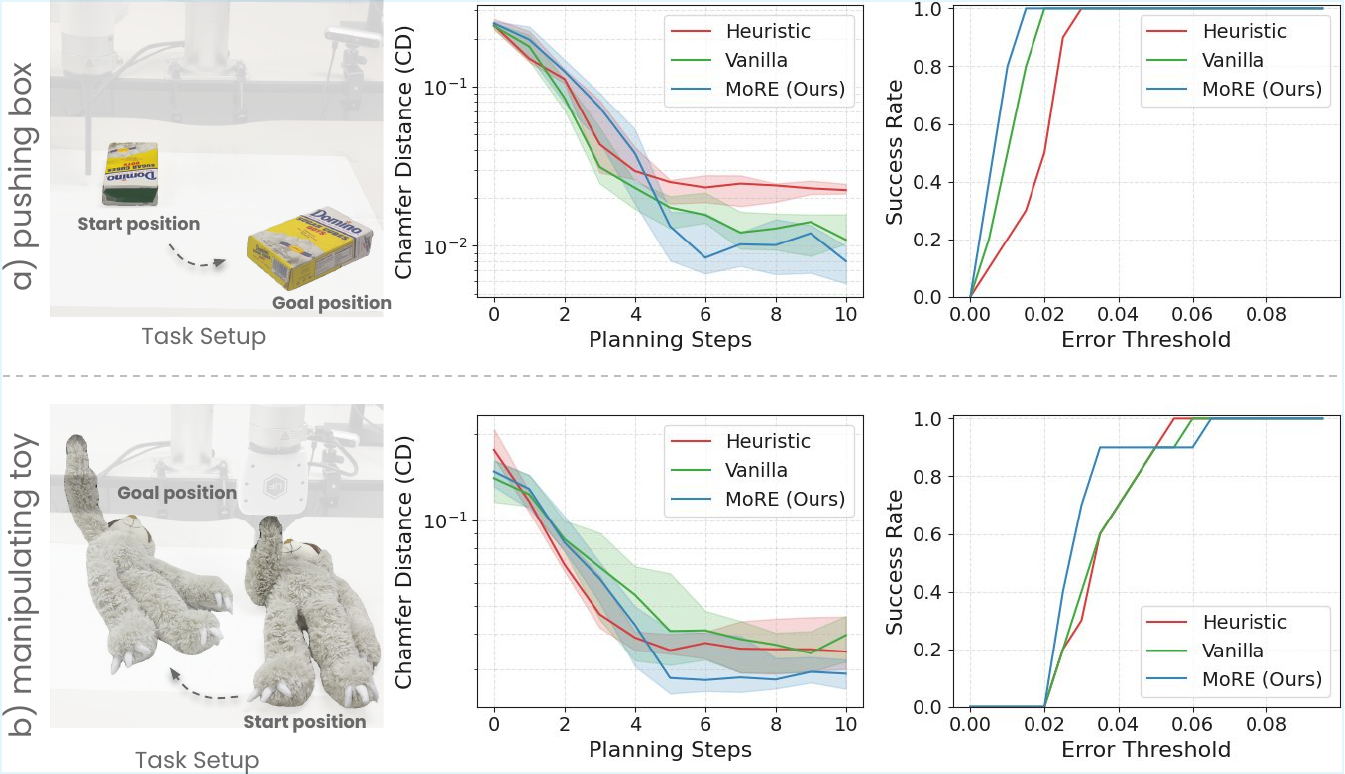}

  \caption{
\textbf{Model Predictive Control (MPC) performance.}
(a) Box pushing and (b) toy manipulation results. Each section details the prediction error across planning steps (left) and success rate vs. error threshold (right). MoRE achieves lower error and higher success than the baselines. 
}
\label{fig:task_mpc_real}

\end{figure*}

%% file: figs/real.tex
\begin{figure*}[t]
  \centering
  \includegraphics[width=\linewidth]{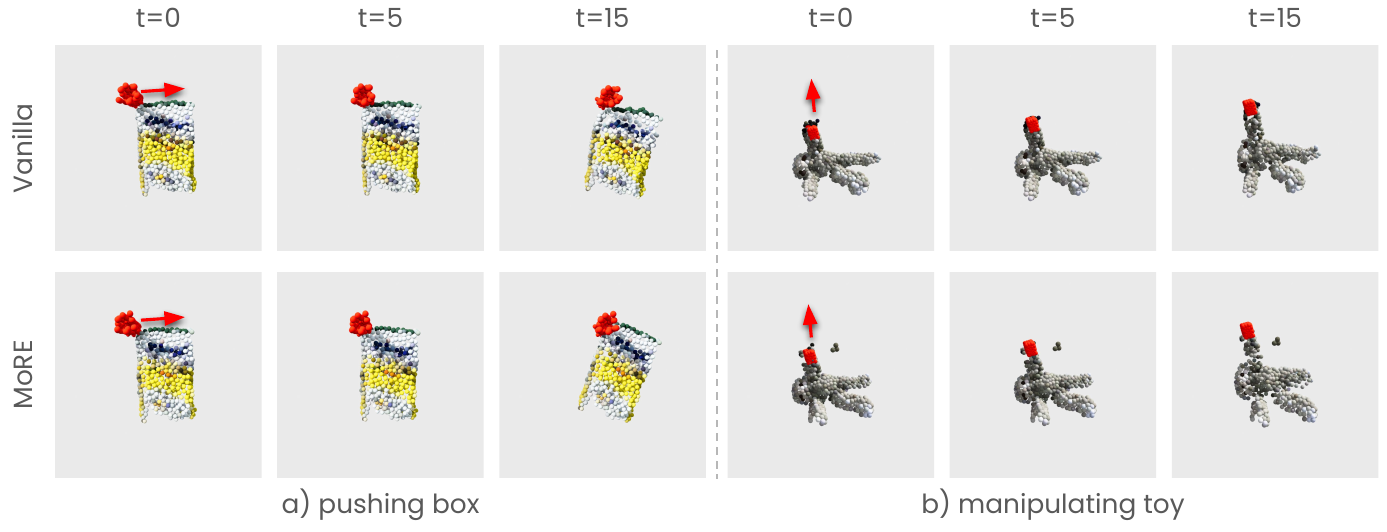}

\caption{
\textbf{Real world dynamics predictions.}  We visualize predictions of point motion on real world data. The red points are the end-effector (action) points. The Vanilla model struggles to treat the box as rigid and the toy sloth as soft. In contrast, our model correctly handles the different cases.
}

\label{fig:real_preds}
\end{figure*}

%% file: 5_conclusion.tex
\section{Discussion and Limitations}
We introduced MoRE, a 3D-aware robotic world model for predicting interaction dynamics across heterogeneous physical regimes. By modeling scenes and actions directly in point cloud space and decomposing dynamics into specialized motion experts, MoRE enables contact-rich reasoning and robust modeling of both rigid and deformable interactions.

\paragraph{Limitations}
Our method currently focuses on modeling the motion of a single manipulated object and does not handle simultaneous multi-object interactions. Extending MoRE to model interactions among multiple objects with potentially different physical behaviors remains an important direction for future work. In addition, MoRE does not explicitly represent fine-grained physical properties, such as mass, friction, and elasticity. This may limit its ability to distinguish visually similar objects with different internal states, such as an empty bottle and a water-filled bottle, whose dynamics can differ despite similar appearances. Incorporating physical property estimation into the world model could improve its ability to predict such interactions and generalize across objects with different physical characteristics.

%% file: supp.tex
\appendix
\section*{Appendix}

The supplementary material provides implementation details of dataset curation, model architecture, MPC pipeline, and more qualitative results of dynamic rollout prediction and real-world results. Project Page: \href{https://alwaysleepy.github.io/MoRE-WM-site/}{MoRE-WM-site}.

\section{Implementation details}
\label{sec:supp-implementation}

\subsection{Dataset Curation} Our training dataset is constructed by mixing simulated and real-world data, with a fixed mixture ratio of 1:1 throughout training. The two data sources are curated under a unified 3D point-cloud representation, enabling the model to benefit from both the scale and controllability of simulation and the realism of real-world observations.

\subsubsection{Simulation Data Generation} The simulated portion is collected using Isaac Lab~\cite{Mittal_Isaac_Lab_-_2025}, leveraging its large-scale parallel simulation capability to efficiently generate diverse and controllable object--tool interaction trajectories. \paragraph{Assets} The simulation dataset includes three categories of assets: rigid objects, deformable objects, and tools. The rigid-object set contains approximately 300 objects spanning 130 categories, sourced from RoboTwin~\cite{chen2025robotwin20scalabledata}. The deformable-object set includes around 300 objects across 100 categories, adapted from~\cite{stojanov21cvpr}. The tool set consists of approximately 30 objects from RoboSmith~\cite{lin2025robotsmith}, functionally categorized into pushing, holding, lifting, and flattening tools, covering a broad range of interaction primitives. \paragraph{Interaction generation} Interaction data collected in simulation focus on non-prehensile manipulation. To ensure diverse and physically plausible interactions, we first uniformly sample contact points on the object surface and use the corresponding surface normal information provided by the simulator to parameterize action directions. Based on this procedure, we construct two primary non-prehensile action types: push, which applies outward forces to displace the object, and reach, which induces inward pulling interactions. This design enables systematic coverage of contact configurations while avoiding hand-crafted action heuristics. 

\paragraph{Data preprocessing} During data collection, we record the full mesh states of both objects and tools at each simulation time step, including rigid-body transformations and per-vertex deformations for soft objects. 
From each mesh, we uniformly sample surface points and apply farthest point sampling (FPS) to obtain point clouds with well-distributed spatial coverage. The number of sampled object points is determined adaptively according to the object size, ranging from 150 to 512 points.
These sampled point clouds provide consistent point correspondences across time and are used to construct object state sequences and tool action sequences as inputs to the world model. 

\paragraph{Domain randomization} To improve generalization and reduce the sim-to-real gap, we apply extensive domain randomization during simulation. Specifically, we randomize physical properties such as mass and friction coefficients for rigid objects, as well as material parameters, such as Young's modulus, for deformable objects. In addition, to model actuation uncertainty commonly observed in real robotic systems, we inject Gaussian noise into each action at every time step.

\input{figs/benchmark}
\input{tables/gating}

\input{figs/supp_sim}

\subsubsection{Real-World Data Processing} 
The real-world portion is curated from two main sources: the PGND dataset~\cite{zhang2025particle} and the DROID dataset~\cite{khazatsky2024droid}.

The PGND dataset provides real object--robot interaction sequences with reconstructed 3D point clouds over time and can be directly incorporated into our training pipeline. 

 For DROID, we follow the PointWorld~\cite{huang2026pointworld} processing pipeline to obtain a dense 3D flow at the scene-level. Since our world model is object-centric, we further process each DROID episode to extract the target object motion. For each episode video, we first use Qwen3~\cite{yang2025qwen3} to identify the end-effector, the object interacted with, and the interaction type. We then apply SAM2~\cite{ravi2025sam} to segment the object specified by the language-based recognition result. When multiple candidate masks are produced, we select the mask closest to the end-effector as the target object. The selected mask is finally used to filter the scene-level 3D flow, yielding the object-centric target flow used for training. 
 

\input{figs/supp_real}

By unifying simulated and real data under a common 3D point-cloud representation and training with a balanced mixture, the model benefits from both the scale and controllability of simulation and the realism and noise characteristics of real-world observations.

\subsection{Model Architecture}
\paragraph{Backbone} The model consists of two main components: a static geometry encoder and a motion encoder with a mixture-of-experts (MoE) design. The static encoder processes the current object point cloud using a frozen Uni3D~\cite{zhou2023uni3d} backbone, which encodes per-point geometric features. Uni3D is instantiated with grouped point tokens and a transformer-based architecture, producing $1024$-dim features; its parameters are fixed during training to provide stable geometric priors. The motion encoder operates on dynamic tokens constructed from object and tool histories, and is implemented as a transformer-based network with multi-head self-attention operating on concatenated object and tool point tokens. The motion encoder outputs per-point $1024$-dim motion features. 

\paragraph{Fusion} We adopt a cross-attention mechanism to fuse static geometric features with motion features, enabling the resulting latent representations to encode dynamic motion information while preserving geometric consistency throughout the interaction process.

\paragraph{Decoding} To decode this latent representation into heterogeneous physical behaviors, we design K different motion experts. Each expert is an MLP head: the rigid expert predicts a global 6D rotation and 3D translation parameterizing an SE(3) transform, while the remaining experts predict per-point 3D displacements to model deformable dynamics. 
Given that different physical regimes exhibit substantially different dynamics patterns, we adopt hard expert routing. Specifically, we design the gating vector as a one-hot vector, which selects a single expert at each time step within an interaction sequence. The gating network is implemented as a two-layer MLP that takes the latent motion features as input and outputs expert selection logits. This design enforces explicit expert specialization and avoids blending incompatible dynamics across physical regimes.

\paragraph{Model forward process} Given a history of object and tool point clouds, the model performs a forward prediction as follows. The current object point cloud is first encoded by the static geometry encoder to obtain per-point geometric features. Object and tool histories are flattened along the temporal dimension and embedded into dynamic tokens, which are processed by the motion encoder to produce motion features. These static and motion features are fused through cross-attention, producing geometry-aware motion representations. The gating network predicts expert weights, and each expert produces a candidate motion prediction. The rigid expert applies a global SE(3) transform to all object points, while deformable experts predict per-point displacements. The final output is selected based on the routing weights from the gating module.

\subsection{MPC Inference Pipeline}
Since expert routing in MoRE depends on historical object dynamics, the world model cannot directly infer the appropriate physical regime from static observations. Therefore, we adopt a two-stage MPC procedure.

\paragraph{Initialization} We first apply a short sequence of random interactions to the object to get observable motion responses. These probing interactions provide the dynamic history required by the gating network, which then predicts the expert corresponding to the underlying physical regime.

\paragraph{MPC procedure} Conditioning on the selected expert, we perform MPC rollouts using the learned world model. For non-prehensile manipulation, we sample 48 candidate push actions by selecting 16 contact points on the object surface and 3 pushing directions per point. For prehensile manipulation, after a grasp is established, we uniformly sample candidate tool displacements within a bounded workspace.
Each candidate is represented as a short “start–end” waypoint trajectory for the tool. For each candidate, we convert the start–end tool motion into a dense action sequence by interpolating the end-effector pose over a fixed horizon. 
Given the current history buffer of object states and actions, we roll out the world model autoregressively for a fixed number of steps. At each step, the predicted object point cloud is appended to the history by shifting the buffer, while the corresponding future action is updated from the precomputed action sequence.
After generating trajectories for all candidates, we compute the goal cost (Chamfer distance) and select the best candidate and the rollout step number (it is often better to stop early rather than complete the full action sequence). The corresponding action is executed and the system replans at the next observation. 

\paragraph{Real-world execution} In real-world experiments we normalize the coordinates by the mean of the object points so that the point clouds match the model’s training distribution. The object point cloud is obtained by fusing point clouds from four Realsense D455 cameras, each masked using SAM3~\cite{carion2025sam3segmentconcepts} with text prompts (``box'', ``sloth''). Statistical outlier removal from Open3D~\cite{Zhou2018} is used to remove points based on average neighbor distance. 
Farthest point sampling is applied in the same way as during training.

\subsection{Gating Network Evaluation}
In this section, we provide a detailed evaluation of the gating network.
The gating network determines which expert is selected and therefore plays a critical role in the accuracy of the final predictions. We evaluate its performance on three categories of tasks: rigid non-prehensile, soft non-prehensile, and soft prehensile manipulation. For each category of tasks, we perform 50 random interaction trials and measure the expert selection accuracy. As shown in Table~\ref{tab:gating}, the gating network can effectively infer the appropriate expert from historical object dynamics, enabling accurate expert routing and high-fidelity dynamics prediction.

\section{Additional Qualitative Results}

\subsection{Benchmark Task Visualization}
We visualize representative samples from the four benchmark tasks used in our evaluation in Fig.~\ref{fig:benchmark}. The tasks span rigid and soft objects under non-prehensile and prehensile interactions across both simulation and real-world settings, covering diverse contact dynamics and interactions.

\subsection{Simulation Visualization}
In this section, we provide additional simulation results on model-based control using world models trained with MoRE. We evaluate our approach on different non-prehensile manipulation tasks involving diverse rigid objects, deformable objects, and various tools. The qualitative results, shown in Figure~\ref{fig:supp_sim}, demonstrate that MoRE enables accurate dynamics prediction, robust handling of heterogeneous physical properties, and generalization across diverse object–tool interactions.

\subsection{Real-world Visualizations}
We visualize real-world model-based control results in Fig.~\ref{fig:supp_real}. Our results show strong model performance under sensor noise and partial observations.

%% file: figs/benchmark.tex
\begin{figure}
  \centering
  \includegraphics[width=0.7\columnwidth]{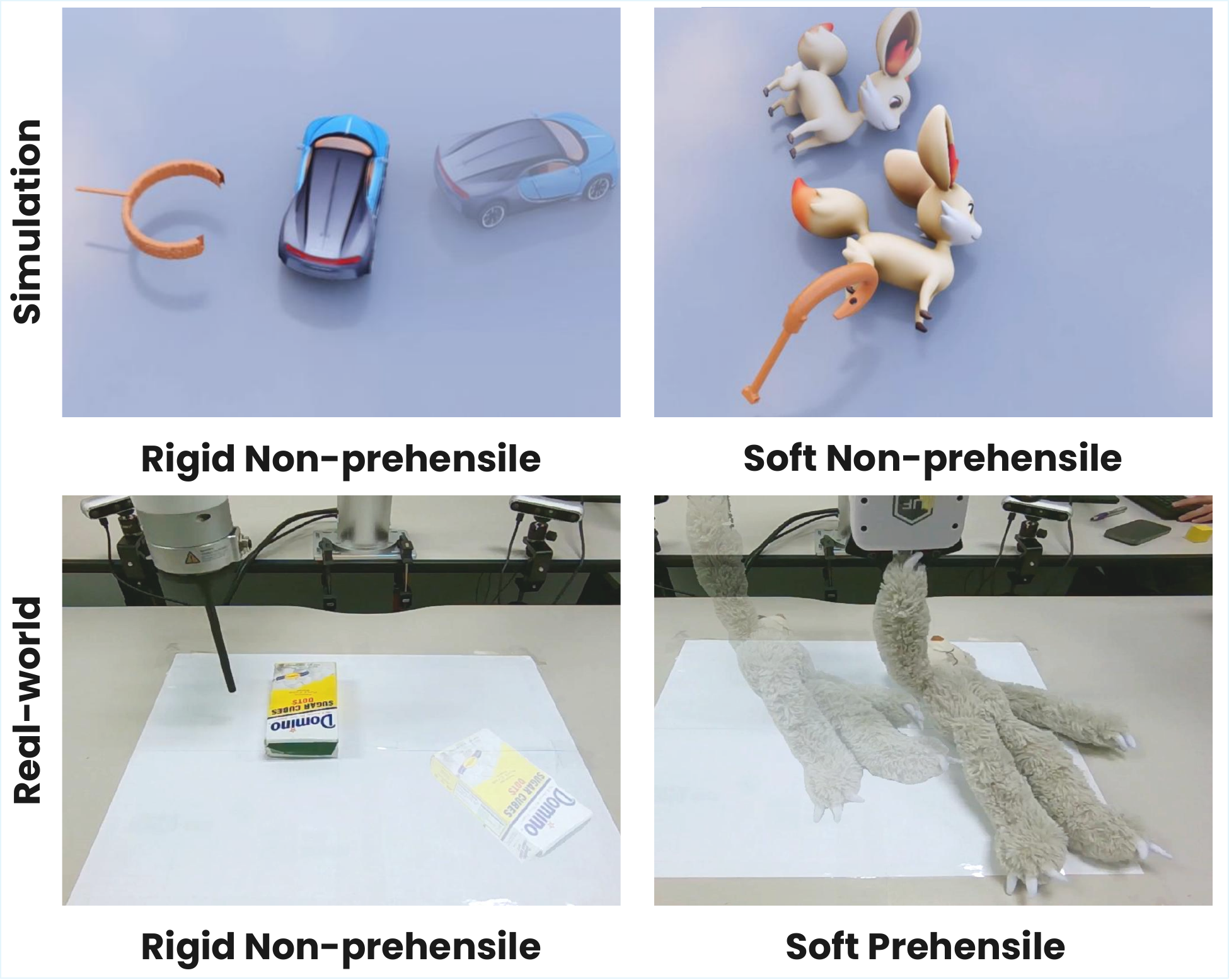}
  \caption{\textbf{Benchmark tasks visualization}: Simulation and real-world benchmarks covering rigid and soft objects under non-prehensile and prehensile interactions, designed to evaluate generalization across diverse physical regimes.}
  \label{fig:benchmark}

\end{figure}

%% file: tables/gating.tex
\begin{table}[t]
\centering
\caption{\textbf{Gating Network Accuracy.}
Expert selection accuracy (\%) evaluated over 50 random interactions per task category. The gating network maintains consistently high accuracy across diverse regimes, indicating reliable expert routing for different physical dynamics.}
\label{tab:gating}
\setlength{\tabcolsep}{10pt}
\renewcommand{\arraystretch}{1.2}
\begin{tabular}{lcccc}
\toprule

& \textbf{Push-X}
& \textbf{Manipulate Toy}
& \textbf{Soft Grasp}
& \textbf{Overall} \\
\midrule
\textbf{Acc.}
& 95.4\% 
& 94.9\% 
& 99.8\% 
& 96.7\% \\
\bottomrule
\end{tabular}
\end{table}

%% file: figs/supp_sim.tex
\begin{figure*}
  \centering
  \begin{subfigure}{\linewidth}
    \centering
    \includegraphics[width=0.99\linewidth]{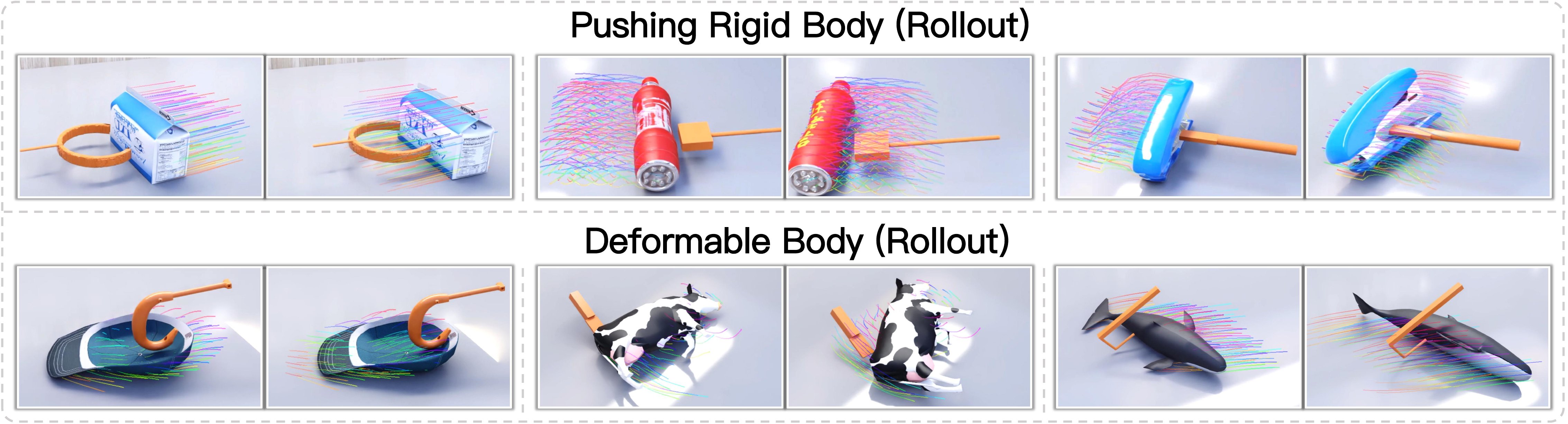}
    \caption{Dynamic prediction in simulation. \textbf{Row 1:} rigid non-prehensile pushing, where the model accurately simulates the rolling behavior of bottles under end-effector contact.
\textbf{Row 2:} deformable object interactions under non-prehensile and prehensile contact, exhibiting both local deformations and global object motion.}
  \end{subfigure}

  \vspace{0.5em}

  \begin{subfigure}{\linewidth}

  \centering
  \includegraphics[width=0.99\linewidth]{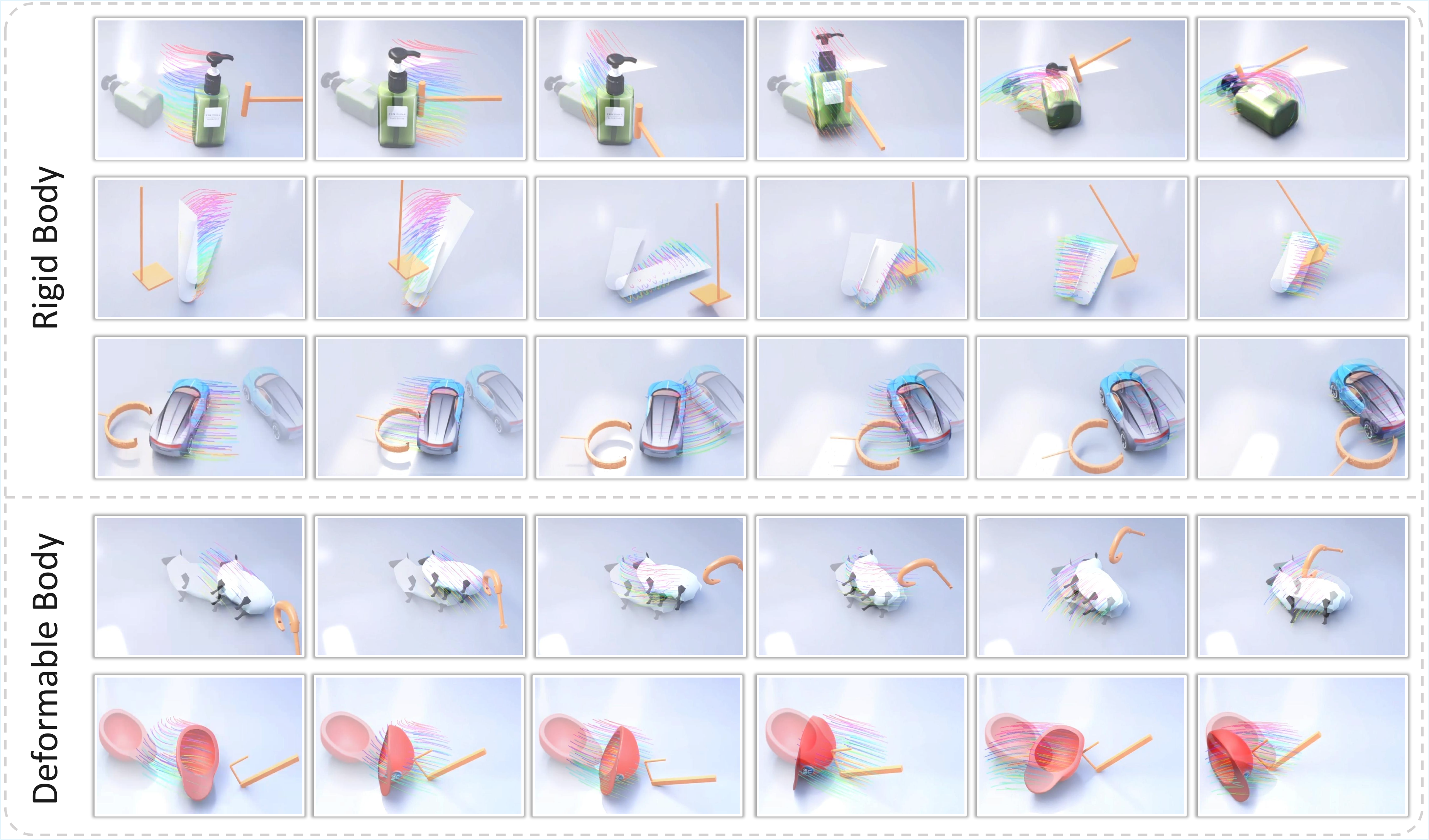}
  \caption{MPC rollout results in simulation tasks. Qualitative simulation results of model-based control using world models trained with MoRE. The results demonstrates robust performance across diverse tools, objects, and physical regimes, including both rigid and deformable interactions.}
  
  \end{subfigure}

  \caption{\textbf{ Simulation results on diverse manipulation tasks.}}
  \label{fig:supp_sim}
\end{figure*}

%% file: figs/supp_real.tex
\begin{figure*}
  \centering
  \begin{subfigure}{\linewidth}
    \centering
    \includegraphics[width=0.99\linewidth]{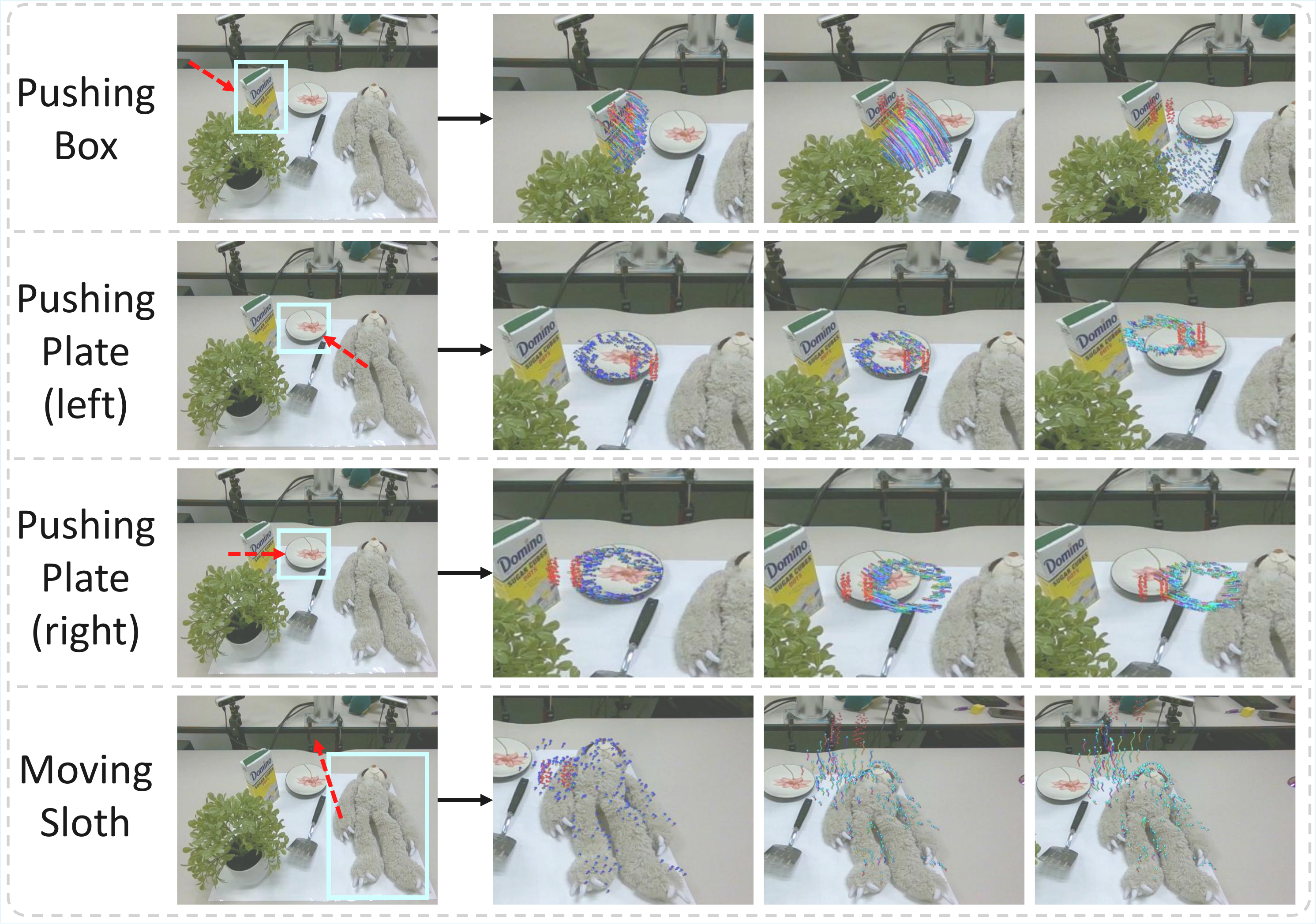}
    \caption{Real-world dynamics prediction under heterogeneous physical properties, demonstrating that our unified model can accurately capture object dynamics across different material regimes.}
  \end{subfigure}

  \vspace{0.5em}

  \begin{subfigure}{\linewidth}
    \centering
    \includegraphics[width=0.99\linewidth]{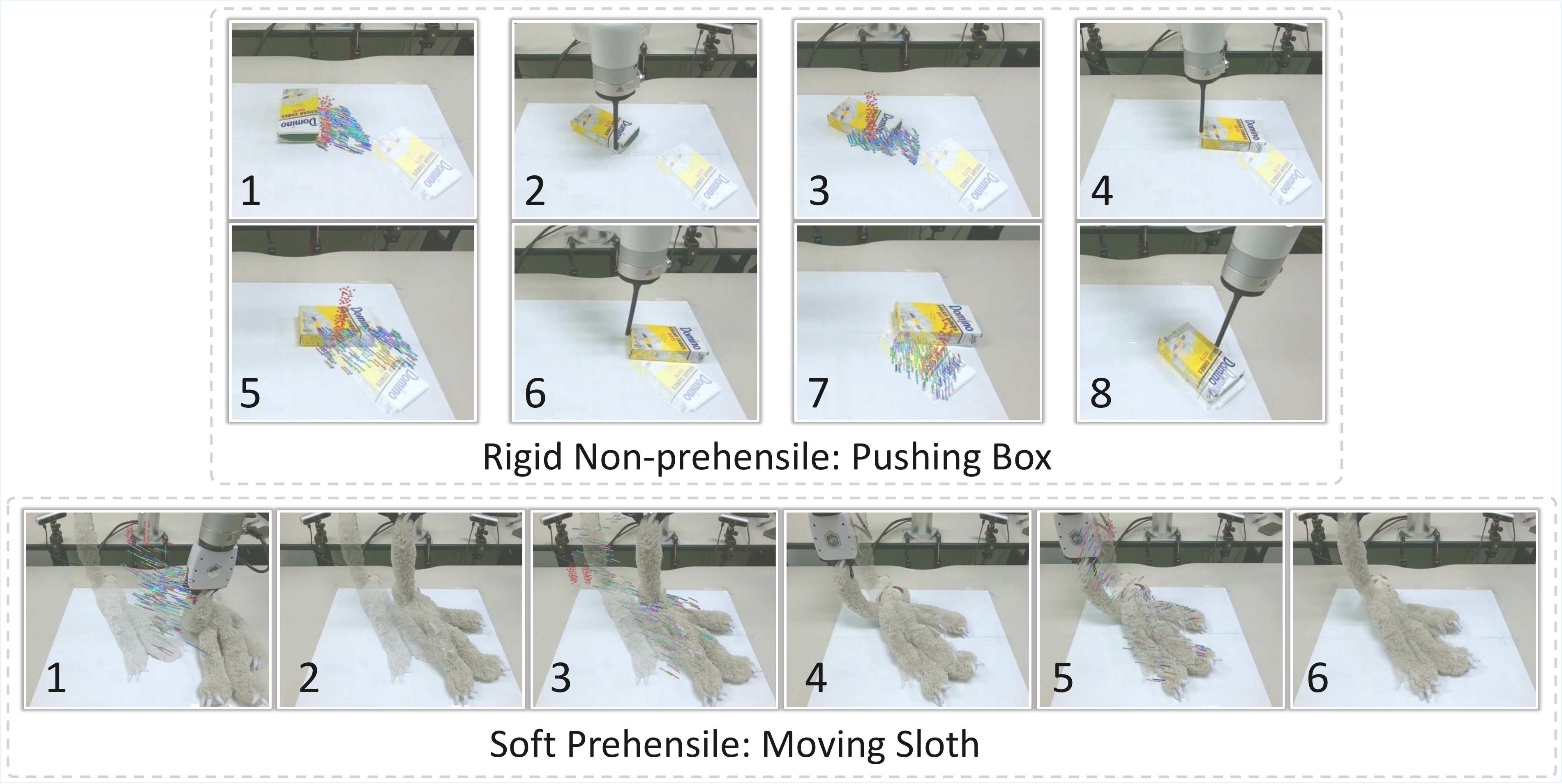}
    \caption{MPC rollouts on real-world manipulation tasks, showing robust control performance under sensor noise and partial observations, and highlighting the model’s ability to handle contact-rich interactions across diverse physical regimes.}
  \end{subfigure}

  \caption{\textbf{ Real-world results on diverse manipulation tasks.}}
  \label{fig:supp_real}
\end{figure*}

%% file: references.bib
@article{zhou2024dino,
  title={Dino-wm: World models on pre-trained visual features enable zero-shot planning},
  author={Zhou, Gaoyue and Pan, Hengkai and LeCun, Yann and Pinto, Lerrel},
  journal={arXiv preprint arXiv:2411.04983},
  year={2024}
}

@article{Zhou2018,
    author    = {Qian-Yi Zhou and Jaesik Park and Vladlen Koltun},
    title     = {{Open3D}: {A} Modern Library for {3D} Data Processing},
    journal   = {arXiv:1801.09847},
    year      = {2018},
}

@misc{lu2026magicsimunifiedinfrastructureexecutable,
      title={MagicSim: A Unified Infrastructure for Executable Embodied Interaction}, 
      author={Haoran Lu and Songling Liu and Yue Chen and Guo Ye and Mutian Shen and Shuyang Yu and Yu Xiao and Shang Wu and Jiayi Wang and Jianshu Zhang and Jihai Zhao and Xiangtian Gui and Chuye Hong and Yuran Wang and Maojiang Su and Ruihai Wu and Zhaoran Wang and Han Liu},
      year={2026},
      eprint={2606.17511},
      archivePrefix={arXiv},
      primaryClass={cs.RO},
      url={https://arxiv.org/abs/2606.17511}, 
}

@misc{chen2025robotwin20scalabledata,
      title={RoboTwin 2.0: A Scalable Data Generator and Benchmark with Strong Domain Randomization for Robust Bimanual Robotic Manipulation}, 
      author={Tianxing Chen and Zanxin Chen and Baijun Chen and Zijian Cai and Yibin Liu and Zixuan Li and Qiwei Liang and Xianliang Lin and Yiheng Ge and Zhenyu Gu and Weiliang Deng and Yubin Guo and Tian Nian and Xuanbing Xie and Qiangyu Chen and Kailun Su and Tianling Xu and Guodong Liu and Mengkang Hu and Huan-ang Gao and Kaixuan Wang and Zhixuan Liang and Yusen Qin and Xiaokang Yang and Ping Luo and Yao Mu},
      year={2025},
      eprint={2506.18088},
      archivePrefix={arXiv},
      primaryClass={cs.RO},
      url={https://arxiv.org/abs/2506.18088}, 
}

@article{le2023differentiable,
  title={Differentiable physics simulation of dynamics-augmented neural objects},
  author={Le Cleac'h, Simon and Yu, Hong-Xing and Guo, Michelle and Howell, Taylor and Gao, Ruohan and Wu, Jiajun and Manchester, Zachary and Schwager, Mac},
  journal={IEEE Robotics and Automation Letters},
  volume={8},
  number={5},
  pages={2780--2787},
  year={2023},
  publisher={IEEE}
}

@inproceedings{chen2022virtual,
  title={Virtual elastic objects},
  author={Chen, Hsiao-yu and Tretschk, Edith and Stuyck, Tuur and Kadlecek, Petr and Kavan, Ladislav and Vouga, Etienne and Lassner, Christoph},
  booktitle={Proceedings of the IEEE/CVF Conference on Computer Vision and Pattern Recognition},
  pages={15827--15837},
  year={2022}
}

@inproceedings{zhang2024physdreamer,
    title={{PhysDreamer}: Physics-Based Interaction with 3D Objects via Video Generation},
    author={Tianyuan Zhang and Hong-Xing Yu and Rundi Wu and Brandon Y. Feng and Changxi Zheng and Noah Snavely and Jiajun Wu and William T. Freeman},
    booktitle={European Conference on Computer Vision},
    year={2024},
    organization={Springer}
    }

@inproceedings{guan2022neurofluid,
  title={Neurofluid: Fluid dynamics grounding with particle-driven neural radiance fields},
  author={Guan, Shanyan and Deng, Huayu and Wang, Yunbo and Yang, Xiaokang},
  booktitle={International conference on machine learning},
  pages={7919--7929},
  year={2022},
  organization={PMLR}
}

@article{shi2024robocraft,
  title={Robocraft: Learning to see, simulate, and shape elasto-plastic objects in 3d with graph networks},
  author={Shi, Haochen and Xu, Huazhe and Huang, Zhiao and Li, Yunzhu and Wu, Jiajun},
  journal={The International Journal of Robotics Research},
  volume={43},
  number={4},
  pages={533--549},
  year={2024},
  publisher={SAGE Publications Sage UK: London, England}
}

@inproceedings{sanchez2020learning,
  title={Learning to simulate complex physics with graph networks},
  author={Sanchez-Gonzalez, Alvaro and Godwin, Jonathan and Pfaff, Tobias and Ying, Rex and Leskovec, Jure and Battaglia, Peter},
  booktitle={International conference on machine learning},
  pages={8459--8468},
  year={2020},
  organization={PMLR}
}

@inproceedings{pfaff2020learning,
  title={Learning mesh-based simulation with graph networks},
  author={Pfaff, Tobias and Fortunato, Meire and Sanchez-Gonzalez, Alvaro and Battaglia, Peter},
  booktitle={International conference on learning representations},
  year={2020}
}

@inproceedings{ma2023learning,
  title={Learning neural constitutive laws from motion observations for generalizable pde dynamics},
  author={Ma, Pingchuan and Chen, Peter Yichen and Deng, Bolei and Tenenbaum, Joshua B and Du, Tao and Gan, Chuang and Matusik, Wojciech},
  booktitle={International Conference on Machine Learning},
  pages={23279--23300},
  year={2023},
  organization={PMLR}
}

@article{ko2023learning,
  title={Learning to act from actionless videos through dense correspondences},
  author={Ko, Po-Chen and Mao, Jiayuan and Du, Yilun and Sun, Shao-Hua and Tenenbaum, Joshua B},
  journal={arXiv preprint arXiv:2310.08576},
  year={2023}
}

@article{hansen2023td,
  title={Td-mpc2: Scalable, robust world models for continuous control},
  author={Hansen, Nicklas and Su, Hao and Wang, Xiaolong},
  journal={arXiv preprint arXiv:2310.16828},
  year={2023}
}

@article{hu2023gaia,
  title={Gaia-1: A generative world model for autonomous driving},
  author={Hu, Anthony and Russell, Lloyd and Yeo, Hudson and Murez, Zak and Fedoseev, George and Kendall, Alex and Shotton, Jamie and Corrado, Gianluca},
  journal={arXiv preprint arXiv:2309.17080},
  year={2023}
}

@inproceedings{lin2022learning,
  title={Learning visible connectivity dynamics for cloth smoothing},
  author={Lin, Xingyu and Wang, Yufei and Huang, Zixuan and Held, David},
  booktitle={Conference on Robot Learning},
  pages={256--266},
  year={2022},
  organization={PMLR}
}

@article{chen2022comphy,
  title={Comphy: Compositional physical reasoning of objects and events from videos},
  author={Chen, Zhenfang and Yi, Kexin and Li, Yunzhu and Ding, Mingyu and Torralba, Antonio and Tenenbaum, Joshua B and Gan, Chuang},
  journal={arXiv preprint arXiv:2205.01089},
  year={2022}
}

@article{ai2025review,
  title={A review of learning-based dynamics models for robotic manipulation},
  author={Ai, Bo and Tian, Stephen and Shi, Haochen and Wang, Yixuan and Pfaff, Tobias and Tan, Cheston and Christensen, Henrik I and Su, Hao and Wu, Jiajun and Li, Yunzhu},
  journal={Science Robotics},
  volume={10},
  number={106},
  pages={eadt1497},
  year={2025},
  publisher={American Association for the Advancement of Science}
}

@article{huang2026pointworld,
  title={PointWorld: Scaling 3D World Models for In-The-Wild Robotic Manipulation},
  author={Huang, Wenlong and Chao, Yu-Wei and Mousavian, Arsalan and Liu, Ming-Yu and Fox, Dieter and Mo, Kaichun and Fei-Fei, Li},
  journal={arXiv preprint arXiv:2601.03782},
  year={2026}
}

@article{zhang2024adaptigraph,
  title={Adaptigraph: Material-adaptive graph-based neural dynamics for robotic manipulation},
  author={Zhang, Kaifeng and Li, Baoyu and Hauser, Kris and Li, Yunzhu},
  journal={arXiv preprint arXiv:2407.07889},
  year={2024}
}

@article{jiang2025phystwin,
  title={Phystwin: Physics-informed reconstruction and simulation of deformable objects from videos},
  author={Jiang, Hanxiao and Hsu, Hao-Yu and Zhang, Kaifeng and Yu, Hsin-Ni and Wang, Shenlong and Li, Yunzhu},
  journal={arXiv preprint arXiv:2503.17973},
  year={2025}
}

@article{huang2025self,
  title={Self Forcing: Bridging the Train-Test Gap in Autoregressive Video Diffusion},
  author={Huang, Xun and Li, Zhengqi and He, Guande and Zhou, Mingyuan and Shechtman, Eli},
  journal={arXiv preprint arXiv:2506.08009},
  year={2025}
}

@inproceedings{huang2025points2plans,
  title={Points2plans: From point clouds to long-horizon plans with composable relational dynamics},
  author={Huang, Yixuan and Agia, Christopher and Wu, Jimmy and Hermans, Tucker and Bohg, Jeannette},
  booktitle={2025 IEEE International Conference on Robotics and Automation (ICRA)},
  pages={1208--1216},
  year={2025},
  organization={IEEE}
}

@article{zhou2024robodreamer,
  title={Robodreamer: Learning compositional world models for robot imagination},
  author={Zhou, Siyuan and Du, Yilun and Chen, Jiaben and Li, Yandong and Yeung, Dit-Yan and Gan, Chuang},
  journal={arXiv preprint arXiv:2404.12377},
  year={2024}
}

@article{huang2025particleformer,
  title={ParticleFormer: A 3D Point Cloud World Model for Multi-Object, Multi-Material Robotic Manipulation},
  author={Huang, Suning and Chen, Qianzhong and Zhang, Xiaohan and Sun, Jiankai and Schwager, Mac},
  journal={arXiv preprint arXiv:2506.23126},
  year={2025}
}

@article{ha2018world,
  title={World models},
  author={Ha, David and Schmidhuber, J{\"u}rgen},
  journal={arXiv preprint arXiv:1803.10122},
  volume={2},
  number={3},
  year={2018}
}

@inproceedings{todorov2012mujoco,
  title={Mujoco: A physics engine for model-based control},
  author={Todorov, Emanuel and Erez, Tom and Tassa, Yuval},
  booktitle={2012 IEEE/RSJ international conference on intelligent robots and systems},
  pages={5026--5033},
  year={2012},
  organization={IEEE}
}

@book{simo1998computational,
  title={Computational inelasticity},
  author={Simo, Juan C and Hughes, Thomas JR},
  year={1998},
  publisher={Springer}
}

@article{liu2023robotic,
  title={Robotic manipulation of deformable rope-like objects using differentiable compliant position-based dynamics},
  author={Liu, Fei and Su, Entong and Lu, Jingpei and Li, Mingen and Yip, Michael C},
  journal={IEEE Robotics and Automation Letters},
  volume={8},
  number={7},
  pages={3964--3971},
  year={2023},
  publisher={IEEE}
}

@article{lan2022affine,
  title={Affine body dynamics: Fast, stable \& intersection-free simulation of stiff materials},
  author={Lan, Lei and Kaufman, Danny M and Li, Minchen and Jiang, Chenfanfu and Yang, Yin},
  journal={arXiv preprint arXiv:2201.10022},
  year={2022}
}

@article{ferguson2021intersection,
  title={Intersection-free rigid body dynamics},
  author={Ferguson, Zachary and Li, Minchen and Schneider, Teseo and Gil-Ureta, Francisca and Langlois, Timothy and Jiang, Chenfanfu and Zorin, Denis and Kaufman, Danny M and Panozzo, Daniele},
  journal={ACM Transactions on Graphics},
  volume={40},
  number={4},
  year={2021}
}

@article{muller2007position,
  title={Position based dynamics},
  author={M{\"u}ller, Matthias and Heidelberger, Bruno and Hennix, Marcus and Ratcliff, John},
  journal={Journal of Visual Communication and Image Representation},
  volume={18},
  number={2},
  pages={109--118},
  year={2007},
  publisher={Elsevier}
}

@article{stomakhin2013material,
  title={A material point method for snow simulation},
  author={Stomakhin, Alexey and Schroeder, Craig and Chai, Lawrence and Teran, Joseph and Selle, Andrew},
  journal={ACM Transactions on Graphics (TOG)},
  volume={32},
  number={4},
  pages={1--10},
  year={2013},
  publisher={ACM New York, NY, USA}
}

@article{hu2018moving,
  title={A moving least squares material point method with displacement discontinuity and two-way rigid body coupling},
  author={Hu, Yuanming and Fang, Yu and Ge, Ziheng and Qu, Ziyin and Zhu, Yixin and Pradhana, Andre and Jiang, Chenfanfu},
  journal={ACM Transactions on Graphics (TOG)},
  volume={37},
  number={4},
  pages={1--14},
  year={2018},
  publisher={ACM New York, NY, USA}
}

@article{de2020material,
  title={Material point method after 25 years: Theory, implementation, and applications},
  author={De Vaucorbeil, Alban and Nguyen, Vinh Phu and Sinaie, Sina and Wu, Jian Ying},
  journal={Advances in applied mechanics},
  volume={53},
  pages={185--398},
  year={2020},
  publisher={Elsevier}
}

@book{commandeur2007introduction,
  title={An introduction to state space time series analysis},
  author={Commandeur, Jacques JF and Koopman, Siem Jan},
  year={2007},
  publisher={Oxford university press}
}

@MISC{coumans2021,
author =   {Erwin Coumans and Yunfei Bai},
title =    {PyBullet, a Python module for physics simulation for games, robotics and machine learning},
howpublished = {\url{http://pybullet.org}},
year = {2016--2021}
}

@article{li2018learning,
  title={Learning particle dynamics for manipulating rigid bodies, deformable objects, and fluids},
  author={Li, Yunzhu and Wu, Jiajun and Tedrake, Russ and Tenenbaum, Joshua B and Torralba, Antonio},
  journal={arXiv preprint arXiv:1810.01566},
  year={2018}
}

@article{zhang2025particle,
  title={Particle-Grid Neural Dynamics for Learning Deformable Object Models from RGB-D Videos},
  author={Zhang, Kaifeng and Li, Baoyu and Hauser, Kris and Li, Yunzhu},
  journal={arXiv preprint arXiv:2506.15680},
  year={2025}
}

@article{baraff2001physically,
  title={Physically based modeling: Rigid body simulation},
  author={Baraff, David},
  journal={Siggraph Course Notes, Acm Siggraph},
  volume={2},
  number={1},
  pages={2--1},
  year={2001}
}

@article{baraff1997introduction,
  title={An introduction to physically based modeling: rigid body simulation I—unconstrained rigid body dynamics},
  author={Baraff, David},
  journal={SIGGRAPH course notes},
  volume={82},
  year={1997}
}

@inproceedings{liu2024physgen,
  title={Physgen: Rigid-body physics-grounded image-to-video generation},
  author={Liu, Shaowei and Ren, Zhongzheng and Gupta, Saurabh and Wang, Shenlong},
  booktitle={European Conference on Computer Vision},
  pages={360--378},
  year={2024},
  organization={Springer}
}

@inproceedings{chen2025physgen3d,
  title={Physgen3d: Crafting a miniature interactive world from a single image},
  author={Chen, Boyuan and Jiang, Hanxiao and Liu, Shaowei and Gupta, Saurabh and Li, Yunzhu and Zhao, Hao and Wang, Shenlong},
  booktitle={Proceedings of the Computer Vision and Pattern Recognition Conference},
  pages={6178--6189},
  year={2025}
}

@article{lin2025robotsmith,
  title={RobotSmith: Generative Robotic Tool Design for Acquisition of Complex Manipulation Skills},
  author={Lin, Chunru and Yuan, Haotian and Wang, Yian and Qiu, Xiaowen and Wang, Tsun-Hsuan and Guo, Minghao and Wang, Bohan and Narang, Yashraj and Fox, Dieter and Gan, Chuang},
  journal={arXiv preprint arXiv:2506.14763},
  year={2025}
}

@misc{carion2025sam3segmentconcepts,
      title={SAM 3: Segment Anything with Concepts},
      author={Nicolas Carion and Laura Gustafson and Yuan-Ting Hu and Shoubhik Debnath and Ronghang Hu and Didac Suris and Chaitanya Ryali and Kalyan Vasudev Alwala and Haitham Khedr and Andrew Huang and Jie Lei and Tengyu Ma and Baishan Guo and Arpit Kalla and Markus Marks and Joseph Greer and Meng Wang and Peize Sun and Roman Rädle and Triantafyllos Afouras and Effrosyni Mavroudi and Katherine Xu and Tsung-Han Wu and Yu Zhou and Liliane Momeni and Rishi Hazra and Shuangrui Ding and Sagar Vaze and Francois Porcher and Feng Li and Siyuan Li and Aishwarya Kamath and Ho Kei Cheng and Piotr Dollár and Nikhila Ravi and Kate Saenko and Pengchuan Zhang and Christoph Feichtenhofer},
      year={2025},
      eprint={2511.16719},
      archivePrefix={arXiv},
      primaryClass={cs.CV},
      url={https://arxiv.org/abs/2511.16719},
}

@inproceedings{stojanov21cvpr,
      title={Using Shape to Categorize: Low-Shot Learning with an Explicit Shape Bias},
      author={Stefan Stojanov and Anh Thai and James M. Rehg},
      booktitle = {Proceedings of the IEEE/CVF Conference on Computer Vision and Pattern Recognition (CVPR)},
      year      = {2021}
}

@inproceedings{zhong2024reconstruction,
  title={Reconstruction and simulation of elastic objects with spring-mass 3d gaussians},
  author={Zhong, Licheng and Yu, Hong-Xing and Wu, Jiajun and Li, Yunzhu},
  booktitle={European Conference on Computer Vision},
  pages={407--423},
  year={2024},
  organization={Springer}
}

@article{qiao2021differentiable,
  title={Differentiable simulation of soft multi-body systems},
  author={Qiao, Yiling and Liang, Junbang and Koltun, Vladlen and Lin, Ming},
  journal={Advances in Neural Information Processing Systems},
  volume={34},
  pages={17123--17135},
  year={2021}
}

@article{du2021diffpd,
  title={Diffpd: Differentiable projective dynamics},
  author={Du, Tao and Wu, Kui and Ma, Pingchuan and Wah, Sebastien and Spielberg, Andrew and Rus, Daniela and Matusik, Wojciech},
  journal={ACM Transactions on Graphics (ToG)},
  volume={41},
  number={2},
  pages={1--21},
  year={2021},
  publisher={ACM New York, NY}
}

@inproceedings{lin2021softgym,
  title={Softgym: Benchmarking deep reinforcement learning for deformable object manipulation},
  author={Lin, Xingyu and Wang, Yufei and Olkin, Jake and Held, David},
  booktitle={Conference on Robot Learning},
  pages={432--448},
  year={2021},
  organization={PMLR}
}

@article{Mittal_Isaac_Lab_-_2025,
author = {Mittal, Mayank and Roth, Pascal and Tigue, James and Richard, Antoine and Zhang, Octi and Du, Peter and Serrano-Muñoz, Antonio and Yao, Xinjie and Zurbrügg, René and Rudin, Nikita and Wawrzyniak, Lukasz and Rakhsha, Milad and Denzler, Alain and Heiden, Eric and Borovicka, Ales and Ahmed, Ossama and Akinola, Iretiayo and Anwar, Abrar and Carlson, Mark T. and Feng, Ji Yuan and Garg, Animesh and Gasoto, Renato and Gulich, Lionel and Guo, Yijie and Gussert, M. and Hansen, Alex and Kulkarni, Mihir and Li, Chenran and Liu, Wei and Makoviychuk, Viktor and Malczyk, Grzegorz and Mazhar, Hammad and Moghani, Masoud and Murali, Adithyavairavan and Noseworthy, Michael and Poddubny, Alexander and Ratliff, Nathan and Rehberg, Welf and Schwarke, Clemens and Singh, Ritvik and Smith, James Latham and Tang, Bingjie and Thaker, Ruchik and Trepte, Matthew and Van Wyk, Karl and Yu, Fangzhou and Millane, Alex and Ramasamy, Vikram and Steiner, Remo and Subramanian, Sangeeta and Volk, Clemens and Chen, CY and Jawale, Neel and Kuruttukulam, Ashwin Varghese and Lin, Michael A. and Mandlekar, Ajay and Patzwaldt, Karsten and Welsh, John and Lafleche, Jean-Francois and Moënne-Loccoz, Nicolas and Park, Soowan and Stepinski, Rob and Van Gelder, Dirk and Amevor, Chris and Carius, Jan and Chang, Jumyung and He Chen, Anka and Ciechomski, Pablo de Heras and Daviet, Gilles and Mohajerani, Mohammad and von Muralt, Julia and Reutskyy, Viktor and Sauter, Michael and Schirm, Simon and Shi, Eric L. and Terdiman, Pierre and Vilella, Kenny and Widmer, Tobias and Yeoman, Gordon and Chen, Tiffany and Grizan, Sergey and Li, Cathy and Li, Lotus and Smith, Connor and Wiltz, Rafael and Alexis, Kostas and Chang, Yan and Fan, Linxi "Jim" and Farshidian, Farbod and Handa, Ankur and Huang, Spencer and Hutter, Marco and Narang, Yashraj and Pouya, Soha and Sheng, Shiwei and Zhu, Yuke and Macklin, Miles and Moravanszky, Adam and Reist, Philipp and Guo, Yunrong and Hoeller, David and State, Gavriel},
doi = {10.48550/arXiv.2511.04831},
journal = {arXiv preprint arXiv:2511.04831},
title = {{Isaac Lab - A GPU-Accelerated Simulation Framework for Multi-Modal Robot Learning}},
url = {https://arxiv.org/abs/2511.04831},
year = {2025}
}

@incollection{sumner2007embedded,
  title={Embedded deformation for shape manipulation},
  author={Sumner, Robert W and Schmid, Johannes and Pauly, Mark},
  booktitle={ACM SIGGRAPH 2007 Papers},
  publisher={Association for Computing Machinery},
  pages={80--es},
  year={2007}
}

@article{zhou2023uni3d,
  title={Uni3d: Exploring unified 3d representation at scale},
  author={Zhou, Junsheng and Wang, Jinsheng and Ma, Baorui and Liu, Yu-Shen and Huang, Tiejun and Wang, Xinlong},
  journal={arXiv preprint arXiv:2310.06773},
  year={2023}
}

@article{di2024keypoint,
  title={Keypoint action tokens enable in-context imitation learning in robotics},
  author={Di Palo, Norman and Johns, Edward},
  journal={arXiv preprint arXiv:2403.19578},
  year={2024}
}

@inproceedings{shridhar2023perceiver,
  title={Perceiver-actor: A multi-task transformer for robotic manipulation},
  author={Shridhar, Mohit and Manuelli, Lucas and Fox, Dieter},
  booktitle={Conference on Robot Learning},
  pages={785--799},
  year={2023},
  organization={PMLR}
}

@article{levine2016end,
  title={End-to-end training of deep visuomotor policies},
  author={Levine, Sergey and Finn, Chelsea and Darrell, Trevor and Abbeel, Pieter},
  journal={Journal of Machine Learning Research},
  volume={17},
  number={39},
  pages={1--40},
  year={2016}
}

@article{shi2023robocook,
  title={Robocook: Long-horizon elasto-plastic object manipulation with diverse tools},
  author={Shi, Haochen and Xu, Huazhe and Clarke, Samuel and Li, Yunzhu and Wu, Jiajun},
  journal={arXiv preprint arXiv:2306.14447},
  year={2023}
}

@inproceedings{wu2023daydreamer,
  title={Daydreamer: World models for physical robot learning},
  author={Wu, Philipp and Escontrela, Alejandro and Hafner, Danijar and Abbeel, Pieter and Goldberg, Ken},
  booktitle={Conference on robot learning},
  pages={2226--2240},
  year={2023},
  organization={PMLR}
}

@article{zheng2025flare,
  title={FLARE: Robot learning with implicit world modeling},
  author={Zheng, Ruijie and Wang, Jing and Reed, Scott and Bjorck, Johan and Fang, Yu and Hu, Fengyuan and Jang, Joel and Kundalia, Kaushil and Lin, Zongyu and Magne, Loic and others},
  journal={arXiv preprint arXiv:2505.15659},
  year={2025}
}

@article{guo2025ctrl,
  title={Ctrl-world: A controllable generative world model for robot manipulation},
  author={Guo, Yanjiang and Shi, Lucy Xiaoyang and Chen, Jianyu and Finn, Chelsea},
  journal={arXiv preprint arXiv:2510.10125},
  year={2025}
}

@article{battaglia2016interaction,
  title={Interaction networks for learning about objects, relations and physics},
  author={Battaglia, Peter and Pascanu, Razvan and Lai, Matthew and Jimenez Rezende, Danilo and others},
  journal={Advances in neural information processing systems},
  volume={29},
  year={2016}
}

@inproceedings{xie2024physgaussian,
  title={Physgaussian: Physics-integrated 3d gaussians for generative dynamics},
  author={Xie, Tianyi and Zong, Zeshun and Qiu, Yuxing and Li, Xuan and Feng, Yutao and Yang, Yin and Jiang, Chenfanfu},
  booktitle={Proceedings of the IEEE/CVF Conference on Computer Vision and Pattern Recognition},
  pages={4389--4398},
  year={2024}
}

@article{fedus2022switch,
  title={Switch transformers: Scaling to trillion parameter models with simple and efficient sparsity},
  author={Fedus, William and Zoph, Barret and Shazeer, Noam},
  journal={Journal of Machine Learning Research},
  volume={23},
  number={120},
  pages={1--39},
  year={2022}
}

@article{shazeer2017outrageously,
  title={Outrageously large neural networks: The sparsely-gated mixture-of-experts layer},
  author={Shazeer, Noam and Mirhoseini, Azalia and Maziarz, Krzysztof and Davis, Andy and Le, Quoc and Hinton, Geoffrey and Dean, Jeff},
  journal={arXiv preprint arXiv:1701.06538},
  year={2017}
}

@article{mendonca2023structured,
  title={Structured world models from human videos},
  author={Mendonca, Russell and Bahl, Shikhar and Pathak, Deepak},
  journal={arXiv preprint arXiv:2308.10901},
  year={2023}
}

@article{robine2023transformer,
  title={Transformer-based world models are happy with 100k interactions},
  author={Robine, Jan and H{\"o}ftmann, Marc and Uelwer, Tobias and Harmeling, Stefan},
  journal={arXiv preprint arXiv:2303.07109},
  year={2023}
}

@article{xu2019densephysnet,
  title={Densephysnet: Learning dense physical object representations via multi-step dynamic interactions},
  author={Xu, Zhenjia and Wu, Jiajun and Zeng, Andy and Tenenbaum, Joshua B and Song, Shuran},
  journal={arXiv preprint arXiv:1906.03853},
  year={2019}
}

@article{oquab2023dinov2,
  title={Dinov2: Learning robust visual features without supervision},
  author={Oquab, Maxime and Darcet, Timoth{\'e}e and Moutakanni, Th{\'e}o and Vo, Huy and Szafraniec, Marc and Khalidov, Vasil and Fernandez, Pierre and Haziza, Daniel and Massa, Francisco and El-Nouby, Alaaeldin and others},
  journal={arXiv preprint arXiv:2304.07193},
  year={2023}
}

@article{bardes2024revisiting,
  title={Revisiting feature prediction for learning visual representations from video},
  author={Bardes, Adrien and Garrido, Quentin and Ponce, Jean and Chen, Xinlei and Rabbat, Michael and LeCun, Yann and Assran, Mahmoud and Ballas, Nicolas},
  journal={arXiv preprint arXiv:2404.08471},
  year={2024}
}

@misc{openai2024video,
  title={Video generation models as world simulators},
  author={OpenAI},
  year={2024}
}

@article{liu2026moright,
  title={MoRight: Motion Control Done Right},
  author={Liu, Shaowei and Ren, Xuanchi and Shen, Tianchang and Ling, Huan and Gupta, Saurabh and Wang, Shenlong and Fidler, Sanja and Gao, Jun},
  journal={arXiv preprint arXiv:2604.07348},
  year={2026}
}

@article{khazatsky2024droid,
  title={Droid: A large-scale in-the-wild robot manipulation dataset},
  author={Khazatsky, Alexander and Pertsch, Karl and Nair, Suraj and Balakrishna, Ashwin and Dasari, Sudeep and Karamcheti, Siddharth and Nasiriany, Soroush and Srirama, Mohan Kumar and Chen, Lawrence Yunliang and Ellis, Kirsty and others},
  journal={arXiv preprint arXiv:2403.12945},
  year={2024}
}

@article{yang2025qwen3,
  title={Qwen3 technical report},
  author={Yang, An and Li, Anfeng and Yang, Baosong and Zhang, Beichen and Hui, Binyuan and Zheng, Bo and Yu, Bowen and Gao, Chang and Huang, Chengen and Lv, Chenxu and others},
  journal={arXiv preprint arXiv:2505.09388},
  year={2025}
}

@inproceedings{ravi2025sam,
  title={Sam 2: Segment anything in images and videos},
  author={Ravi, Nikhila and Gabeur, Valentin and Hu, Yuan-Ting and Hu, Ronghang and Ryali, Chaitanya and Ma, Tengyu and Khedr, Haitham and R{\"a}dle, Roman and Rolland, Chloe and Gustafson, Laura and others},
  booktitle={International Conference on Learning Representations},
  volume={2025},
  pages={28085--28128},
  year={2025}
}
